\documentclass[10pt,twocolumn,letterpaper]{article}

\usepackage[pagenumbers]{cvpr} 

\usepackage{multirow}
\usepackage{array}
\newcolumntype{L}[1]{>{\raggedright\arraybackslash}p{#1}}
\usepackage{makecell}
\usepackage{colortbl,xcolor,booktabs}

\newcommand{\name}{LLaMo\xspace}

\usepackage{tcolorbox}
\newtcolorbox{promptbox}{
  colback=gray!10,       
  colframe=gray!50,      
  rounded corners,
  boxrule=0.8pt,
  arc=4mm,               
  left=10pt, right=10pt,
  top=8pt, bottom=8pt,
  fontupper=\ttfamily\small  
}

\definecolor{cvprblue}{rgb}{0.21,0.49,0.74}
\usepackage[pagebackref,breaklinks,colorlinks,allcolors=cvprblue]{hyperref}

\def\paperID{3264} 
\def\confName{CVPR}
\def\confYear{2026}

\title{\name: Scaling Pretrained Language Models for Unified Motion Understanding and Generation with Continuous Autoregressive Tokens}

\author{Zekun Li$^{1,2*}$ \quad Sizhe An$^{2}$ \quad Chengcheng Tang$^{2}$ \quad Chuan Guo$^{2}$ \quad Ivan Shugurov$^{2}$ \quad Linguang Zhang$^{2}$ \\ Amy Zhao$^{2}$ \quad Srinath Sridhar$^{1}$ \quad Lingling Tao$^{2}$ \quad Abhay Mittal$^{2}$  \\
$^{1}$Brown University \quad $^{2}$Meta\\
}

\begin{document}
\newcommand{\teaserCaption}{
}
\twocolumn[{
    \renewcommand\twocolumn[1][]{#1}
    \maketitle
    \centering
    \begin{minipage}{0.98\textwidth}
        \centering 
        \includegraphics[width=\textwidth]{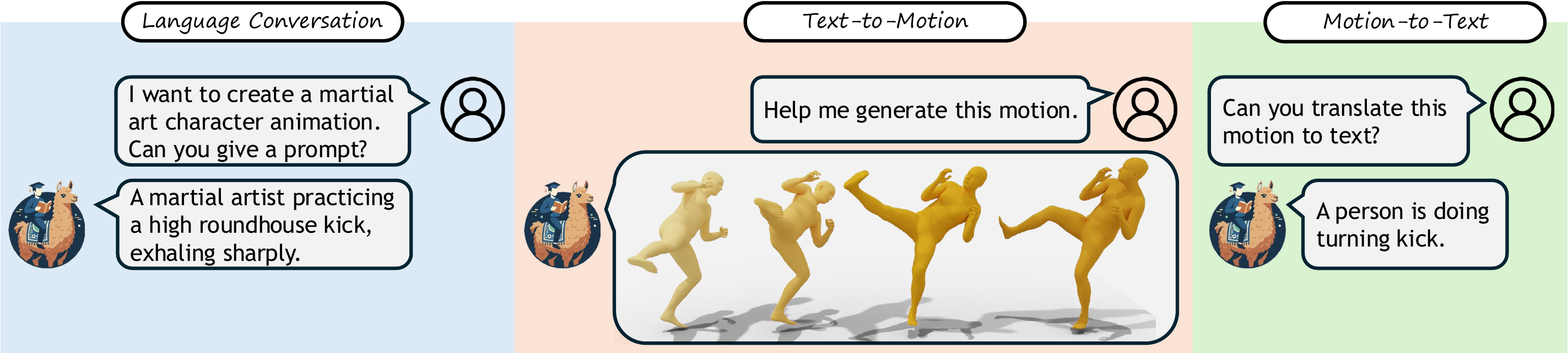}
    \end{minipage}
    
    \captionsetup{type=figure}
    \vspace{-2mm}
    \captionof{figure}{\teaserCaption We introduce \name, the first large-scale motion-language model supporting unified motion understanding and generation without compromising the language proficiency of the underlying LLM.
    }
    \label{fig:teaser}
    \vspace{5mm}
}]

\maketitle
\begingroup
\renewcommand{\thefootnote}{}
\footnotetext{$^{*}$ Work performed during an internship at Meta}
\endgroup

\begin{abstract}
Recent progress in large models has led to significant advances in unified multimodal generation and understanding. However, the development of models that unify motion-language generation and understanding remains largely underexplored. Existing approaches often fine-tune large language models (LLMs) on paired motion-text data, which can result in catastrophic forgetting of linguistic capabilities due to the limited scale of available text-motion pairs.
Furthermore, prior methods typically convert motion into discrete representations via quantization to integrate with language models, introducing substantial jitter artifacts from discrete tokenization.
To address these challenges, we propose \name, a unified framework that extends pretrained LLMs through a modality-specific Mixture-of-Transformers (MoT) architecture. This design inherently preserves the language understanding of the base model while enabling scalable multimodal adaptation.
We encode human motion into a causal continuous latent space and maintain the next-token prediction paradigm in the decoder-only backbone through a lightweight flow-matching head, allowing for streaming motion generation in real-time ($\ge$30 FPS).
Leveraging the comprehensive language understanding of pretrained LLMs and large-scale motion-text pretraining, our experiments demonstrate that \name achieves high-fidelity text-to-motion generation and motion-to-text captioning in general settings, especially zero-shot motion generation, marking a significant step towards a general unified motion-language large model.
Here are our project homepage: \href{https://kunkun0w0.github.io/project/LLaMo/}{https://kunkun0w0.github.io/project/LLaMo/}.
\end{abstract}    
\section{Introduction}
The field of unified multimodal understanding and generation models (UMMs) has recently garnered substantial attention across image~\cite{deng2025emerging,team2025nextstep, ma2025janusflow, xie2024show}, video~\cite{xie2025show,tan2025omni,wei2025univideo}, and audio~\cite{xu2025qwen2,xie2025x, xu2025qwen3omni} modalities.
By integrating both understanding and generation within an end-to-end framework, UMMs enable bidirectional multimodal interaction.This allows the models not only to interpret, but also to produce modality-consistent content with semantic consistency, contextual grounding, and generalization~\cite{team2024chameleon,chen2025janus,deng2025emerging,cui2025emu3}. 
This superior capability relies on large-scale paired multimodal datasets to achieve cross-modal alignment, as well as massive text-only corpora to preserve or enhance language understanding and reasoning abilities~\cite{team2025nextstep,deng2025emerging,wu2025janus,ma2025janusflow,chen2025janus,cui2025emu3,wang2024emu3}.


However, these requirements pose particular challenges for building large-scale human motion–language models, as high-quality paired motion-text data (\eg Mocap data) is much scarcer and more expensive to obtain compared to other modalities such as images and videos.
Nevertheless, directly fine-tuning the text parameters of LLMs with only text-motion data leads to catastrophic forgetting of language abilities~\cite{shi2024lmfusion}, leading to a significant drop in text performance~\cite{xie2025show,hu2025hmvlm,shi2024lmfusion}. 
This degradation undermines the reasoning potential of large UMMs during the post-training stage~\cite{guo2025deepseek,wang2025unirl}, where preserving strong language competence is crucial for UMMs to maintain coherent cross-modal reasoning capabilities~\cite{yue2025does}.
This pivotal capability is necessary for a wide range of downstream motion tasks~\cite{jiang2025solami,zhang2025social,zhao2025navigating}, including prompt refinement~\cite{Wang_2025_ICCV,ouyang2025motion} and multimodal conversational modeling~\cite{jiang2024motionchain,Fang_2025_CVPR}.

Another challenge in building unified motion–language models lies in the tokenization of motion data. 
Existing unified motion–language models either discretize motion through quantization ~\cite{jiang2023motiongpt,fan2025go, wang2024motiongpt} or use continuous tokens but lose the ability to autoregressively model arbitrarily long sequences~\cite{zhu2025motiongpt3}. 
Given the inherently continuous and variable-length nature of human motion, both approaches are suboptimal. 
Discretization introduces jitter-related artifacts~\cite{cho2025discord}, while fixed-length generation mechanisms restrict the model to synthesize motions with a predetermined duration.
This limitation is unrealistic for human motion, where different motion types span diverse temporal scales and require flexibility to accommodate real-world variability. 

These challenges motivate an alternative paradigm: \textit{Can we extend existing pretrained LLMs with the unified capability of understanding and autoregressively generating high-fidelity human motion, while preserving their frontier text-only performance?}

Therefore, we introduce \name, a framework that endows pretrained LLMs with the ability to understand and generate 3D human motion while preventing catastrophic forgetting of text-only performance.
\name achieves this through several key design choices:
(1)~\name adopts a modality-specific Mixture-of-Transformers (MoT) architecture (see \cref{fig:pipeline}), which models separate motion and language parameters while enabling cross-modal communication through shared self-attention. 
By freezing the text-related modules and updating the motion-specific parameters only, we effectively preserve the linguistic competence of the pretrained LLM. 
(2)~To enable high fidelity motion generation of arbitrary length, \name represents human motion in a continuous causal latent space and models the next-token distribution for autoregressive modeling through a flow-matching head~\cite{lipman2022flow,team2025nextstep,li2024autoregressive}.
The continuous latent space is constructed using a causal temporal variational autoencoder, which compactly encodes motion sequences in a streaming manner with high temporal downsampling rate facilitating real-time generation.
Supporting a continuous motion representation allows \name to avoid quantization artifacts and preserve high-frequency micro-dynamics and semantics essential for holistic motion understanding and generation.
\textit{With some optimizations, our large model can achieve real-time streaming motion generation.}

\begin{figure}
    \centering
        \includegraphics[width=1\linewidth]{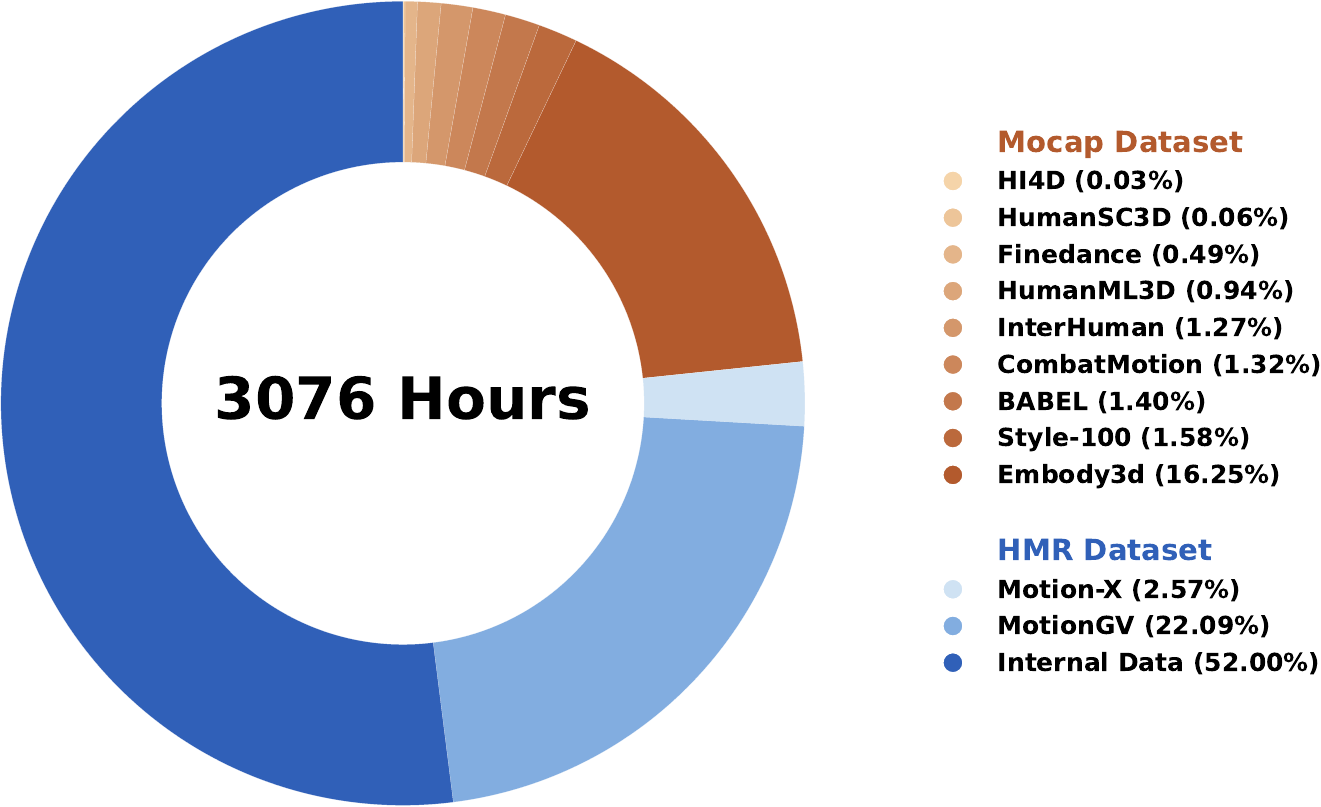}
    \caption{\textbf{Dataset Composition.} We gather a large-scale human motion dataset by combining high quality Mocap datasets with large-scale HMR estimated datasets.}
    \vspace{-0.2cm}
    \label{fig:dataset}
\end{figure}

Finally, to achieve generalizable motion–language understanding and generation, we conduct large-scale pretraining on a newly built in-house dataset containing \textit{over 3 millions motion sequences (3,076 hours)}, composed of Mocap data and human mesh recovery (HMR) estimated motion from human-centric videos, as shown in~\cref{fig:dataset}. 
We evaluate the performance of our model on standard text-to-motion and motion-to-text evaluation protocols established in prior works~\cite{guo2022generating,guo2022tm2t}.
Although HumanML3D~\cite{guo2022generating} comprises less than 1\% of our training data, our model still achieves performance comparable to existing methods trained directly on HumanML3D for both text-to-motion generation and motion-to-text understanding with various SOTA methods.
We further compare our results against the recent large-scale text-to-motion method~\cite{fan2025go} on HumanML3D and show competitive performance. 
To validate the generalization of our model, we also following the MotionMillion-Eval~\cite{fan2025go} to evaluate the zero-shot motion generation capability.

Overall, our contributions can be summarized as follows:
\begin{itemize}
    \item We propose \name, a generic framework to extend pretrained LLMs for human motion generation and understanding, while preserving the original text-only performance via a modality-specific Mixture-of-Transformers (MoT) architecture.
    \item \name~encodes 3D human motion in a causal continuous latent space and employs flow matching to bridge discrete text prediction and continuous motion synthesis, eliminating quantization loss and enabling smooth, dynamic, and text-aligned real-time streaming motion generation.
    \item Comprehensive quantitative and qualitative results demonstrate high fidelity motion generation and faithful motion understanding across various settings.
\end{itemize}
To our knowledge, \name is the first framework to extend pretrained LLMs for unified motion-language modeling while preserving native text performance. 
\section{Related Works}

\paragraph{Architectural design of Unified Multimodal Models.}
The success of decoder-only Transformer~\cite{vaswani2017attention} architectures in large language models (LLMs)~\cite{brown2020language,touvron2023llama} has inspired extensive efforts to extend the language-modeling paradigm to multimodal domains. 
Early work focused on task-specific models (generation vs understanding), using modality-specific encoders for understanding~\cite{pmlr-v202-li23q,liu2023visual} and latent encoders for generation~\cite{ramesh2021zero}.

More recently, unified models for multimodal understanding and generation have gained significant attention~\cite{team2024chameleon,tian2024mm,wang2024emu3,wu2025janus,lu2022unified,chen2025janus,xie2025show,team2025nextstep,zhou2024transfusion,zhao2024monoformer,xie2024show,ma2025janusflow}. 
Unified Multimodal Models (UMMs) generally fall into two main categories:(1) Autoregressive discrete token models, which maintain token-wise prediction for multimodal generation and understanding~\cite{team2024chameleon,wang2024emu3,wu2025janus,chen2025janus,tian2024mm,lu2022unified};
(2) Hybrid Autoregressive-Diffusion Models, which fuse discrete next-token prediction for text with continuous diffusion-based generation for other modalities, such as images, within a single transformer using well-designed attention masks~\cite{xie2024show,deng2025emerging,cui2025emu3,liao2025mogao,shi2024lmfusion,zhou2024transfusion}

While these two UMM designs have been widely validated, they are limited in their ability to support continuous token generation and flexible-length context generation.
To enable streaming generation with a continuous motion codec, we use a flow-matching head that samples continuous motion latents from the autoregressive backbone, following the design of~\cite{team2025nextstep,li2024autoregressive}.
To preserve the language capability of the pretrained LLM, we adopt a modality-specific Mixture-of-Transformers (MoT)~\cite{shi2024lmfusion} design and freeze the text-related modules, preserving linguistic competence of the base LLM during multimodal adaptation.
\vspace{-1em}
\paragraph{Unified Multimodal Human Motion Models.}
Recent years have seen a surge of interest in multimodal motion generation~\cite{lin2023motion,tevet2022human,guo2024momask,lu2025scamo,xu2025mospa,jiang2024scaling,xu2024inter,Meng_2025_CVPR,zhang2025egoreact}.
Most approaches employ a pretrained text encoder, using its embeddings as conditioning signals for motion generation models.
These models can be autoregressive~\cite{fan2025go,xiao2025motionstreamer,zhang2023generating} or generate entire motion sequences in one go~\cite{guo2024momask, tevet2022human, zhang2023remodiffuse}. 

To achieve semantically aligned and contextually grounded motion generation and understanding, several works have explored unified human motion modeling~\cite{wang2024motiongpt,zhu2025motiongpt3,jiang2023motiongpt}.
These methods typically fine-tune pretrained LLMs to support motion generation and understanding, either through full weight training~\cite{jiang2023motiongpt,cao2025being} or parameter-efficient approaches~\cite{wang2024motiongpt,wu2024motion}, and rely on discrete motion codebooks via vector quantization in a non-causal manner.
A recent work~\cite{zhu2025motiongpt3} introduced a MoT-based approach with continuous motion latents, similar to our design.
However, it neither preserves the language capability of the base LLM nor supports streaming motion generation, as it generates motion of fixed length by padding a predetermined number of ($\mathrm{<motion\_out>}$) tokens as Transformer inputs in a single forward pass.
Furthermore, its use of a non-causal motion VAE further restricts the model’s ability to generate motion autoregressively, limiting its applicability to streaming and interactive scenarios.

In contrast to our work, all of these methods finetune the text parameters of the original LLM which leads to a severe drop in the language modeling performance~\cite{hu2025hmvlm}. 
To our knowledge, our work represents the first attempt to integrate human motion into foundational LLMs without hurting their native language performance, supporting streaming motion generation in real-time.  
\section{Method}
In this section we introduce our unified large motion-language model, \name.
First, we describe the motion representation and motion tokenization process, where human motion sequences are converted into continuous tokens using a causal Variational Autoencoder (VAE).
Next, we present the architectural design of our unified motion LLM, highlighting the mixture-of-transformers design that preserves language modality information and the next-token prediction mechanism based on flow matching \etc. 
Finally, we detail our training strategy, including dataset curation and multi-stage optimization framework used to train our model effectively.

\subsection{Motion Representation}
We follow previous works ~\cite{xiao2025motionstreamer,fan2025go} to adopt a 272-dim motion representation, which helps mitigate errors introduced by the inverse kinematics process in the HumanML3D format~\cite{guo2022generating} 
while preserving redundant information (\eg joint location and velocity).
Specifically, it is defined as a tuple comprising:
\begin{equation}
    m_i = \{\dot{r}^x, \dot{r}^z, \dot{r}^a, p^i, v^i,r^i\} \quad ,
\end{equation}
where $(\dot{r}^x, \dot{r}^z)\in\mathbb{R}$ are the root linear velocity on the ground plane, $\dot{r}^a \in \mathbb{R}^6$ is the 6D rotations~\cite{zhou2019continuity} for root angular velocity, $p^i\in\mathbb{R}^{3N}$ is the  local joint positions, $v^i\in\mathbb{R}^{3N}$ is the local joint linear velocities, $r^i\in\mathbb{R}^{6N}$ is the local rotations, and $N$ denotes the number of joints.

\begin{figure*}[ht]
    \centering
    \includegraphics[width=1\linewidth]{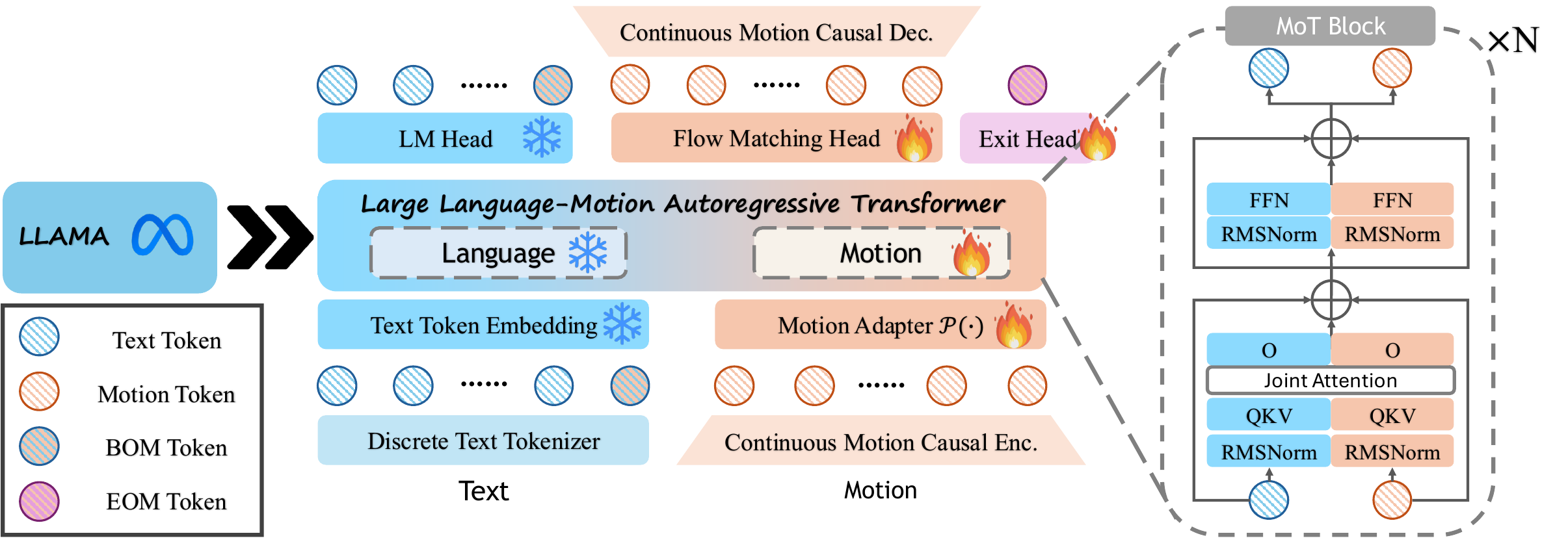}
    \caption{\textbf{Framework overview of \name.} We utilize modality-specific Mixture-of-Transformer (MoT) to process text and motion tokens separately, while enabling cross-modal interactions through shared self-attention.
    To preserve the language performance of the base model, text-related modules are frozen.
    The $[\mathrm{BOM}]$ and $[\mathrm{EOM}]$ tokens denote the start and end of the motion sequence, respectively. An additional exit head allows the model to support flexible-length motion generation.}
    \label{fig:pipeline}
\end{figure*}
\subsection{Continuous Motion Tokenization}
Unlike most previous motion-language models~\cite{cao2025being,fan2025go,jiang2023motiongpt} that rely on discrete motion tokenization and thereby suffer from quantization errors, we encode motion sequence $\{m_i\}_{1:N}$ into a causal continuous latent space.
Specifically, we use a causal CNN-based causal VAE~\cite{xiao2025motionstreamer}, which reconstructs motion frames while strictly preserving temporal causality throughout the sequence.
Given a motion sequence $\{m_1, m_2, ..., m_N\}$, we use the encoder $\mathrm{Enc}_{\phi}$ to model the distribution of motion latent as a set of temporal Gaussian distribution parameters $\{(\mu_1, \sigma_1^2), (\mu_2, \sigma_2^2), ..., (\mu_{N//l}, \sigma_{N//l}^2)\}$ with $z_i\in \mathcal{N}(\mu_i,\sigma_i^2)$, where $l$ represents the temporal downsampling rate of $\mathrm{Enc}_{\phi}$.
We follow the training objective in~\cite{xiao2025motionstreamer}: 
\begin{equation}
    \mathcal{L} = \mathcal{L}_{\mathrm{recon}} + D_\mathrm{KL}\big(\mathrm{Enc}_{\phi}(z|m)\|p(z)\big) + \lambda_\mathrm{root} \mathcal{L}_{\mathrm{root}},
\end{equation}
where $p(z)=\mathcal{N}(0,\mathbf{I})$, $\mathcal{L}_{\mathrm{recon}}$ is the motion representation reconstruction loss, and $\mathcal{L}_{\mathrm{root}}$ is the root representation reconstruction loss.

Although the continuous motion codec can achieve high-fidelity reconstruction from the causal VAE latent space, posterior collapse during VAE reconstruction learning causes instability in generation training and results in a fragile autoregressive behavior in next-token prediction~\cite{sun2024multimodal,shao2025continuous}. 
Unlike discrete autoregressive modeling, where token sampling through softmax inherently tolerates probabilistic noise, flow matching sampling in continuous autoregressive generation operates in a dense latent space where even minor deviations in sampled latents may accumulate and propagate through subsequent steps~\cite{team2025nextstep,ke2025hyperspherical}.
Consequently, the latent decoder must be highly robust to sampling imperfections from the flow-matching head, ensuring stability and fidelity in motion synthesis.
To this end, instead of predicting the variance of the latent distribution as in a traditional VAE, we manually sample the variance from a uniform distribution~\cite{sun2024multimodal} to obtain a robust causal VAE, shown as \cref{eq:sigma_vae}, where $C_{\sigma}=0.01$. We share more details on this in the Appendix.
\begin{equation}
    \begin{aligned}
\mu &= \mathrm{Enc}_{\phi}(m) \\
z &= \mu + \sigma \odot \epsilon, \, \text{where } \epsilon \sim \mathcal{N}(0,\mathbf{I}), \; \sigma \sim \mathbf{U}(0, C_{\sigma}) \\
\hat{m} &= \mathrm{Dec}_{\psi}(z)
\end{aligned}
    \label{eq:sigma_vae}
\end{equation}

\subsection{Unified Motion-Language Model Architecture}
In this section, we present the key design of \name, which extends pretrained LLMs with unified motion generation and understanding capabilities through continuous token autoregressive modeling, while preserving the original language performance.
Our model is built upon decoder-only transformer architecture of Llama \cite{touvron2023llama} as shown in~\cref{fig:pipeline}.
\paragraph{Modality-Specific Mixture-of-Transformers.}
By leveraging MoT blocks, we separate the parameters according to the modality of input tokens, while still facilitating cross-modal interactions through shared self-attention.
Given the input token embeddings $h$, the next layer output embedding $h^\prime$ is formulated as follows, where $h[i]$ means the position index in input multimodal embedding sequence.
\begin{align*}
    &h_{\text{in}}=\begin{cases}
        \text{\colorbox{cyan!10}{\text{RMSNorm\(_\text{T}\)}}}(h[i]),& \, \text{if } h[i] \text{ is text} \\
        \text{\colorbox{orange!30}{\text{RMSNorm\(_\text{M}\)}}}(h[i]),& \, \text{if } h[i]\text{ is motion}
    \end{cases} \\
    &h_\text{Q}, h_\text{K},h_\text{V} = \begin{cases}
        \text{\colorbox{cyan!10}{\text{QKV\(_\text{T}\)}}}(h_{\text{in}}[i]),& \, \text{if }h[i] \text{ is text} \\
        \text{\colorbox{orange!30}{\text{QKV\(_\text{M}\)}}}(h_{\text{in}}[i]),& \, \text{if }h[i]\text{ is motion}
    \end{cases} \\
    &h_\text{O} = \begin{cases}
        \text{\colorbox{cyan!10}{\text{O\(_\text{T}\)}}}\big(\text{Attn}(h_\text{Q}, h_\text{K},h_\text{V})[i]\big),& \,\text{if } h[i] \text{ is text} \\
        \text{\colorbox{orange!30}{\text{O\(_\text{M}\)}}}\big(\text{Attn}(h_\text{Q}, h_\text{K},h_\text{V})[i]\big),& \,\text{if } h[i]\text{ is motion}
    \end{cases} \\
    &h_{\text{mid}}=h_\text{O}+h \\
    &h_{\text{MLP}}=\begin{cases}
        \text{\colorbox{cyan!10}{\text{RMSNorm\(_\text{T}\)}}}(h_\text{mid}[i]),& \, \text{if }h[i] \text{ is text} \\
        \text{\colorbox{orange!30}{\text{RMSNorm\(_\text{M}\)}}}(h_\text{mid}[i]),& \, \text{if }h[i]\text{ is motion}
    \end{cases} \\
    &h^\prime=\begin{cases}
        \text{\colorbox{cyan!10}{\text{FFN\(_\text{T}\)}}}(h_\text{O}[i])+h_\text{mid}[i],& \, \text{if }h[i] \text{ is text} \\
        \text{\colorbox{orange!30}{\text{FFN\(_\text{M}\)}}}(h_\text{O}[i])+h_\text{mid}[i],& \,\text{if } h[i]\text{ is motion}
    \end{cases}
\end{align*}
where $\text{O}(\cdot)$ is the output MLP of attention~\cite{vaswani2017attention}. 
This modality-disentangled design separates network parameters into modality-specific groups, enabling extension of pretrained LLMs to new modalities while preserving base model performance by freezing existing modules.
This approach is model-agnostic, enabling extension of any large language model with motion capabilities without degrading language performance.

\paragraph{Unified Motion-Language Embeddings.}
To process different modality inputs by the unified auto-regressive backbone, 
we adopt a motion adapter $\mathcal{P}(\cdot)$ to align the motion VAE latent space with the language embedding space. 
We structure the text embeddings $x^\mathrm{text}$ and motion embeddings $x^\mathrm{motion}=\mathcal{P}(z)$ based on motion VAE latent $z$ into a sequence following a general interleaved QA format similar to MotionGPT~\cite{jiang2023motiongpt}:
\begin{center}
[BOS] \{Text\} [BOM] \{Motion\} [EOM] \{Text\} $\cdots$ [EOS],
\end{center}
where [BOM] and [EOM] are the special text tokens represent the boundary of the input motion embeddings in the interleaved multimodal input embedding sequence.
To simulate the training-inference gap in token distribution during autoregressive modeling, we follow~\cite{pasini2024continuous} and add random noise  $\eta\in\mathcal{N}(0,0.01)$ on our input motion VAE latent $z$ when we use teacher forcing to train our UMM in motion generation instruction tuning tasks.

\paragraph{Discrete Language Decoding Head.}
We preserve the original sampling mechanism in the base LLM.
Let $\hat{h}[i]^\mathrm{text}$ denote the $i$-th last-layer hidden state of the transformer decoder in the output sequence, $x[i]^\mathrm{text}$ is the $i$-th embedding in the input sequence which represents text modality, and $W_\mathrm{text}$ is the LM head embedding. 
The distribution for computing $x[i]^\mathrm{text}$ is modeled as follows:
\begin{equation}
    P\Big(x[i]^\mathrm{text}\Big| x[{<i}]\Big) = \mathrm{softmax}(\hat{h}[i]^\mathrm{text}W_\mathrm{text})
\end{equation}
During motion understanding tasks, we use the next-token prediction objective to encourage the model output correct text token corresponding to the motion caption.
\begin{equation}
    \mathcal{L}_\mathrm{NTP} = -\mathbb{E}_{x[i]\in\mathrm{text}} \Big[\log P\Big(x[i]\Big| x[{<i}]\Big)\Big]
\end{equation}

\paragraph{Continuous Motion Decoding Head.}
We model the next-motion-token distribution for a given auto-regressive motion last layer hidden state output $\hat{h}[i]^\mathrm{motion}$ using flow matching~\cite{lipman2022flow}.
Specifically, we adopt a light-weight flow matching head $f_\theta(\cdot)$~\cite{team2025nextstep,li2024autoregressive} to predict the defined velocity $v_t=\frac{\mathrm{d}{x_t}}{\mathrm{d}t}$ using $\hat{h}_i^\mathrm{motion}$ as the classifier-free guidance condition.
Let $x_0=z$ denote a clean motion VAE latent, random noise $\epsilon \sim \mathcal{N}(0,\mathbf{I})$, and timestep $t\in[0,1]$, we define the forward process using rectified flow interpolation~\cite{liu2022flow}: $x_t = (1-t)\epsilon+tx_0$.
The velocity field $v_t=x_0-\epsilon$ represents the optimal transport path.
The learning objective for the flow head
can be formalized as: 
\begin{equation}
\mathcal{L}_\mathrm{FM}=\mathbb{E}_{t\in[0,1]} \| f(x_t, t, \hat{h}_i^\mathrm{motion}) - v_t(x) \|
\end{equation}
To stabilize the flow matching training, we resample the timestep $t$ by $k=4$ times for any given $\hat{h}_i^\mathrm{motion}$, since the condition distribution is shifting during the training.

\paragraph{Motion Generation Exit Head.}
Since \name uses continuous motion latents rather than discrete motion tokens, it cannot rely on the traditional strategy to end the autoregressive generation, \ie terminate the motion generation when end of motion token [EOM] appeared.
To address this, following the approach used in TransformerTTS~\cite{li2019neural} and SpeechT5~\cite{ao2022speecht5}, we introduce a binary classifier with a fully connected layer to the output of the decoder-only transformer, and compute the binary cross-entropy loss, $\mathcal{L}_\mathrm{End}$, for motion generation ending signal prediction.
We provide more details in the Appendix.

\subsection{Training Recipe}
In this section, we show the all the training configuration and data curation of our large model pretraining.
\paragraph{Dataset Composition}
In order to learn robust motion-language alignment for our unified multimodal motion model, we gather a large-scale motion-text dataset for training.
Our dataset construction process integrates human motion reconstruction from large-scale in-the-wild video sources and the re-aggregation of existing motion datasets. 
To enhance diversity and coverage, we incorporate multiple established datasets, including HumanML3D~\cite{guo2022generating}, Motion-X~\cite{lin2023motion}, 100-Style~\cite{mason2022real}, CombatMotion~\cite{CombatMotion}, MotionGV~\cite{fan2025go}, InterHuman~\cite{liang2024intergen}, BABEL~\cite{punnakkal2021babel}, FineDance~\cite{li2023finedance}, HI4D~\cite{yin2023hi4d}, HumanSC3D~\cite{fieraru2021learning}, and Embody3d~\cite{mclean2025embody}.

We further scale up our dataset by leveraging an in-house human-centric video dataset.
We extract 3D human motion using GVHMR~\cite{shen2024world} and preprocess the motion representation following MotionMillion~\cite{fan2025go}.
To get textual captions, we directly utilize Gemini-2.5Pro~\cite{comanici2025gemini} to get a diverse set of motion prompts from the videos, since there are serious hallucinations in the MotionMillion~\cite{fan2025go} motion captions during the LLM rewrite stage.
The details of motion dataset curation pipeline can be found in Appendix.

\paragraph{Multi-stage Training.}
\name is trained based on the following objective:
\begin{equation}
    \mathcal{L} = \mathcal{L}_\mathrm{FM} + \lambda_1 \mathcal{L}_\mathrm{NTP} + \lambda_2 \mathcal{L}_\mathrm{End}.
\end{equation}
To effectively train this large model with different modalities and objectives, we present a training recipe that facilitiates stable optimization and cross-modal alignment, summarized in ~\cref{tab:training_recipe}. \\
$\blacktriangleright$ \textbf{Stage 1 (Feature Alignment)}:
Embeddings from different modalities vary in scale and distribution, which can cause training instability~\cite{liu2023visual,xie2025show}.
We first train the motion adapter $\mathcal{P}(\cdot)$ together with the flow matching head to align feature representations across modalities.
This stage aligns motion with the LLM representation space, which stabilizes training and improves convergence later.
\\
$\blacktriangleright$ \textbf{Stage 2 (Joint Learning of AR and FM)}: 
Subsequently, we train the full model, excluding the causal motion VAE and text-related parameters, using the entire motion-text paired dataset.
During training, we observe that the flow-matching head tends to exhibit loss spikes, while the motion understanding objective converges much faster and can easily dominate the optimization of the motion branch.
To mitigate this imbalance, we: (i) reduce the sampling rate of motion-to-text data, following ~\cite{deng2025emerging,team2025nextstep} and (ii) sample four time steps per motion token when training the flow-matching head in the text-to-motion task, following ~\cite{li2024autoregressive}.
Additionally, distinct learning-rate schedules are applied across modules to further stabilize joint training. \\
$\blacktriangleright$ \textbf{Stage 3 (Motion Head Annealing)}:
Finally, we refine the motion prediction head and exit head to improve the output quality while keeping all the other model parameters frozen.
This stage stabilizes optimization, mitigates the instability observed in joint training, and leads to improved synthesis quality.
To further enhance latent stability when generating expressive motions with large dynamics, we filter out under-expressive samples, particularly from MotionGV \cite{fan2025go} and our internal dataset.
The filtering details are provided in the Appendix.

\begin{table}[ht]
    \centering
    \setlength{\tabcolsep}{4pt}
    \small
    
    \begin{tabular}{lccc}
    \toprule
    \textbf{Hyperparameters} & \textbf{Stage 1} & \textbf{Stage 2} & \textbf{Stage 3} \\
    \midrule
    Base LR & $1\times10^{-4}$ & $1\times10^{-4}$ & $1\times10^{-5}$ \\
    AR LR Scheduler & Constant & Cosine & - \\
    Head LR Scheduler & Constant & Constant & Cosine \\
    CE Weight $\lambda_1$ & 0.05 & 0.05 & 0 \\
    BCE Weight $\lambda_2$ & 1e-3 & 1e-3 & 1e-2 \\
    Training Steps & 100K & 200K & 50K \\
    \midrule
    \textbf{Task Ratio} & & & \\
    Text-to-Motion & 0.5 & 0.8 & 1 \\
    Motion-to-Text & 0.5 & 0.2 & 0 \\
    \midrule
    \makecell{\textbf{Trainable Module} \\ \textbf{(No VAE)}} & \makecell{Projector \\ Flow Head} & \makecell{Full Model \\ (w/o Text Params.)} & \makecell{Flow Head \\ Exit Head}\\
    \bottomrule
    \end{tabular}
    \vspace{-0.2cm}
    \caption{\textbf{Training recipe.} We adopt a three-stage training strategy to stabilize our large model training, each focusing on different aspects of model optimization.}
    \label{tab:training_recipe}
\end{table}

\section{Experiments}
\subsection{Motion Reconstruction Evaluation}
\paragraph{Evaluation Metrics.}
To demonstrate the high fidelity of motion codec (causal VAE), we adopt the following metrics: (1) Mean Per Joint Position Error (MPJPE) and Mean Per Joint Rotation Error (MPJRE), which measure the average distance between predicted and ground-truth joint positions and rotations
(2) Symmetric Jerk Percentage Error (sJPE)~\cite{cho2025discord}: a, which assesses under-reconstructed motions and frame-level noise via jerk;
(3) Compression (Comp.): the storage ratio of the motion latent to the input motion representation.
\begin{table}[ht]
    \centering
    \setlength{\tabcolsep}{2pt}
    \small
    \resizebox{0.99\linewidth}{!}{
    \begin{tabular}{lcccccc}
    \toprule
     & MPJPE $\downarrow$ & MPJRE $\downarrow$ & sJPE $\downarrow$ & T.Down& Comp.$\downarrow$\\
    \midrule
    Real Motion  & 0.0 & 0.0 & 0.0 &1& $\times$100\% \\
    \midrule
    FSQ-z512-c64000~\cite{fan2025go} & 41.9 & 6.31 & 0.710 & 2&$\times$94.1\%\\
    CausalTAE-z16 & 32.3 & 6.07 & 0.738 & 4 & $\times$1.47\% \\
    CausalTAE-z32  & 10.1 & 2.58 & 0.586 & 4&$\times$2.94\% \\
    CausalTAE-z64  & 3.86 & 0.68 & 0.389  & 4&$\times$5.88\% \\
    \bottomrule
    \end{tabular}
    }
    \vspace{-0.2cm}
    \caption{\textbf{Motion Tokenization.} We compared the SOTA discrete motion tokenization solution~\cite{fan2025go} with our continuous causal motion tokenization, where `z' means the latent feature dimension and `c' denotes the size of discrete codebook. T.Down denotes the temporal downsampling rate of the motion encoder.}
    \label{tab:tae_comp}
\end{table}

\paragraph{Motion Codec Comparison.}
We compare our continuous causal motion VAE based with the quantization-based FSQ-VAE from MotionMillion~\cite{fan2025go}.
As shown in~\cref{tab:tae_comp}, FSQ requires a large codebook (64k entries) and a high-dimensional embedding (512 dim), but yields low-fidelity reconstruction.
Due to the limited representational capacity of FSQ, further increasing the temporal downsampling rate in the motion encoder becomes challenging.
A higher downsampling rate would require each quantized token to represent a longer motion segment, demanding greater expressive power from the codebook. However, with a finite number of discrete codes, FSQ struggles to capture the fine-grained temporal variations within these extended segments.
As a result, the achievable downsampling rate directly constrains how much the motion token sequence can be shortened, which is an important factor influencing the efficiency of the framework during both training and inference.
In contrast, our continuous causal motion VAE compresses the input motion into compact latent vectors with ease.
We choose a latent dimensionality of $z=32$, as higher-dimensional latent spaces tend to introduce instability when training the MLP-based flow-matching head~\cite{li2024autoregressive}.
\vspace{-0.2cm}
\subsection{Quantitative Results}
In this section, we benchmark our large-scale unified motion–language model on HumanML3D~\cite{guo2022generating}, evaluating both text-to-motion generation and motion-to-text captioning, \textit{even though this dataset contributes less than 1\% of our training data}. 
Due to limited space, we provide additional experiments (e.g. zero-shot evaluation, training recipe ablation) and analysis on our results in the appendix. 
\vspace{-1em}
\paragraph{Text-to-Motion Generation.}
We compare \name not only with large-scale text-to-motion models trained on million-level datasets~\cite{fan2025go}, but also with existing specialist models~\cite{tevet2022human,chen2023executing,zhang2023generating,jiang2023motiongpt,guo2024momask,zhong2023attt2m,xiao2025motionstreamer} that are specifically trained on the HumanML3D dataset~\cite{guo2022generating}, as shown in~\cref{tab:t2m_hl3d}, following the evaluation protocol in~\cite{xiao2025motionstreamer}.
Although \cite{fan2025go} identifies a substantial semantic distribution gap between HumanML3D and large-scale motion corpora, scaling enables both MotionMillion and our model to generate human motions on HumanML3D that remain semantically coherent according to competitive R-precision.
However, due to the limited scale and poor generalization of HumanML3D, the FID metric becomes unreliable and it fails to reflect true motion quality and instead largely captures the dataset gap.
In our experiments, we also observe the emerging phenomenon reported in~\cite{fan2025go}, where generation performance significantly improves as the model scales from 1B to 3B.
Benefiting from the deeply fused text conditioning in MoT and its advanced language understanding capabilities, our model is more robust to rare textual inputs and can stably generate human motions that are better aligned with the intended semantics.
\begin{table}[ht]
    \centering
    \setlength{\tabcolsep}{2pt}
    \small
    \resizebox{0.98\linewidth}{!}{
    \begin{tabular}{lcccccc}
    \toprule
    Methods & FID $\downarrow$ & R@1 $\uparrow$ & R@2 $\uparrow$ & R@3 $\uparrow$ & MM-D $\downarrow$ & Div $\rightarrow$ \\
    \midrule
    HumanML3D~\cite{guo2022generating} & - & 0.702 & 0.864 & 0.914 & 15.151 & 27.492 \\
    \midrule
    \rowcolor{gray!15}\multicolumn{7}{l}{\makebox[\dimexpr\linewidth-2\tabcolsep\relax][l]{\textbf{Only Train on HumanML3D}}} \\
    MDM~\cite{tevet2022human} & 23.454 & 0.523 & 0.692 & 0.764 & 17.423 & 26.325 \\
    MLD~\cite{chen2023executing} & 18.236 & 0.546 & 0.730 & 0.792 & 16.638 & 26.352 \\
    T2M-GPT~\cite{zhang2023generating} & 12.475 & 0.606 & 0.774 & 0.838 & 16.812 & {27.275} \\
    MotionGPT~\cite{jiang2023motiongpt} & 14.375 & 0.456 & 0.598 & 0.628 & 17.892 & 27.114 \\                                                        MoMask~\cite{guo2024momask} & \underline{12.232} & \underline{0.621} & \underline{0.784} & \underline{0.846} & 16.138 & 27.127 \\
    AttT2M~\cite{zhong2023attt2m} & 15.428 & 0.592 & 0.765 & 0.834 & \textbf{15.726} & 26.674 \\
    MotionStreamer~\cite{xiao2025motionstreamer} & \textbf{11.790} & \textbf{0.631} & \textbf{0.802} & \textbf{0.859} & \underline{16.081} & \underline{27.284} \\
    \midrule
    \rowcolor{gray!15}\multicolumn{7}{l}{\makebox[\dimexpr\linewidth-2\tabcolsep\relax][l]{\textbf{Train Large Scale Dataset (HumanML3D is round 1\% of the data)}}} \\
    MotionMillion-3B~\cite{fan2025go} & 23.755 & 0.602 & 0.749 & 0.817 & 16.995 & 26.634\\
    MotionMillion-7B~\cite{fan2025go} & 23.582 & 0.616 & 0.752 & 0.819 & 16.938 & 26.829\\
    \name-1B (our) & 53.942 & 0.541 & 0.689 & 0.761 & 18.215 & 26.846 \\
    \name-3B (our) & 22.491 & 0.606 & 0.766 & 0.839 & 17.057 & \textbf{27.582} \\
    \bottomrule
    \end{tabular}}
    \vspace{-0.2cm}
    \caption{\textbf{Text-to-Motion on HumanML3D.} We compared methods with different training settings, following the evaluation in~\cite{xiao2025motionstreamer}. Our results show comparable metrics to both MotionMillion~\cite{fan2025go} and specialist models.}
    \label{tab:t2m_hl3d}
\end{table}

\vspace{-1.5em}
\paragraph{Motion-to-Text Caption.}
We follow~\cite{guo2022tm2t} protocols to evaluate motion-to-text captioning on HumanML3D~\cite{guo2022generating}, comparing with~\cite{chuan2022tm2t,jiang2023motiongpt,li2024lamp,wu2024mote,zhu2025motiongpt3}. 
To our knowledge, no prior work trains motion understanding on large-scale datasets.
Furthermore, \name is the only work which does not fine tune the text parameters of the underlying LLM. 
As shown in~\cref{tab:m2t_hl3d}, our superior CIDEr~\cite{vedantam2015cider} performance highlights the strong key information captioning performance of our model.
And the competitive BERTScore~\cite{zhang2019bertscore} demonstrates our generated captions have high similarity with ground truth in contextual meaning at the sentence level. 
Different from CIDEr and BERTScore, BLEU~\cite{papineni2002bleu} and ROUGE~\cite{lin2004rouge} both rely on n-gram overlap, making them highly sensitive to exact word choice and surface phrasing.
The lower BLEU@1 but still good BLEU@4 and ROUGE scores indicate diverse or natural wording benefited from large-scale motion-text dataset and advanced language capability. 
Overall, these metrics indicate that our method achieves precise alignment between human motion and text within a unified model.
However, unlike the clear gains observed in motion generation, scaling model size does not yield similar improvements for the understanding task. 

    
    
    
    
\begin{table}[ht]
    \centering
    \resizebox{1\linewidth}{!}{
    \begin{tabular}{lccccc}
  \toprule
     Methods
     & Bleu@1 $\uparrow$ & Bleu@4 $\uparrow$ 
     & Rouge $\uparrow$ & Cider $\uparrow$ & BertScore $\uparrow$ \\
  \midrule
    Real    & 100    & 100    & 100    & 120    & 100    \\
    \midrule 
    \rowcolor{gray!15}\multicolumn{6}{l}{\makebox[\dimexpr\linewidth-2\tabcolsep\relax][l]{\textbf{Only Train on HumanML3D}}} \\
    
    TM2T~\citep{chuan2022tm2t} & \underline{48.9} & 7.00 & {38.1} & 16.8 & 32.2 \\
    
    MotionGPT~\citep{jiang2023motiongpt}  & 48.2 & {12.47} & 37.4 & {29.2} & {32.4} \\
    
    LaMPM2T~\citep{li2024lamp} & 47.8 & \underline{13.04} & 37.1 & 28.9 & {32.7}\\
    MoTe~\citep{wu2024mote} & 46.7 & 11.15 & 37.4 & 31.5 & 30.3 \\
    MotionGPT3~\cite{zhu2025motiongpt3}   & \textbf{59.1} & \textbf{19.41} & \textbf{46.2} & 28.7 & \textbf{35.2}
    \\
    \midrule
    \rowcolor{gray!15}\multicolumn{6}{l}{\makebox[\dimexpr\linewidth-2\tabcolsep\relax][l]{\textbf{Train Large Scale Dataset (HumanML3D is round 1\% of the data)}}} \\
    \name-1B (our) & 36.7 & 10.68 & 38.4 & \textbf{104.7} & 33.3 \\
    \name-3B (our) & 38.3 & 12.06 & \underline{39.9} & \underline{100.8} & \underline{34.8} \\
    
 \bottomrule
 \end{tabular}
 } 
    \vspace{-0.2cm}
    \caption{\textbf{Motion-to-Text on HumanML3D} follow~\cite{guo2022tm2t} protocols. Our results demonstrate competitive performance with other specialist models without optimizing text parameters.}
    \label{tab:m2t_hl3d}
\end{table}

\subsection{Zero-shot Text-to-Motion Qualitative Results}
\begin{figure*}[ht]
    \centering

    \begin{subfigure}[b]{0.32\textwidth}
        \centering
        \begin{minipage}[t][4.2cm][t]{\textwidth}  
            \centering
            \includegraphics[width=\linewidth]{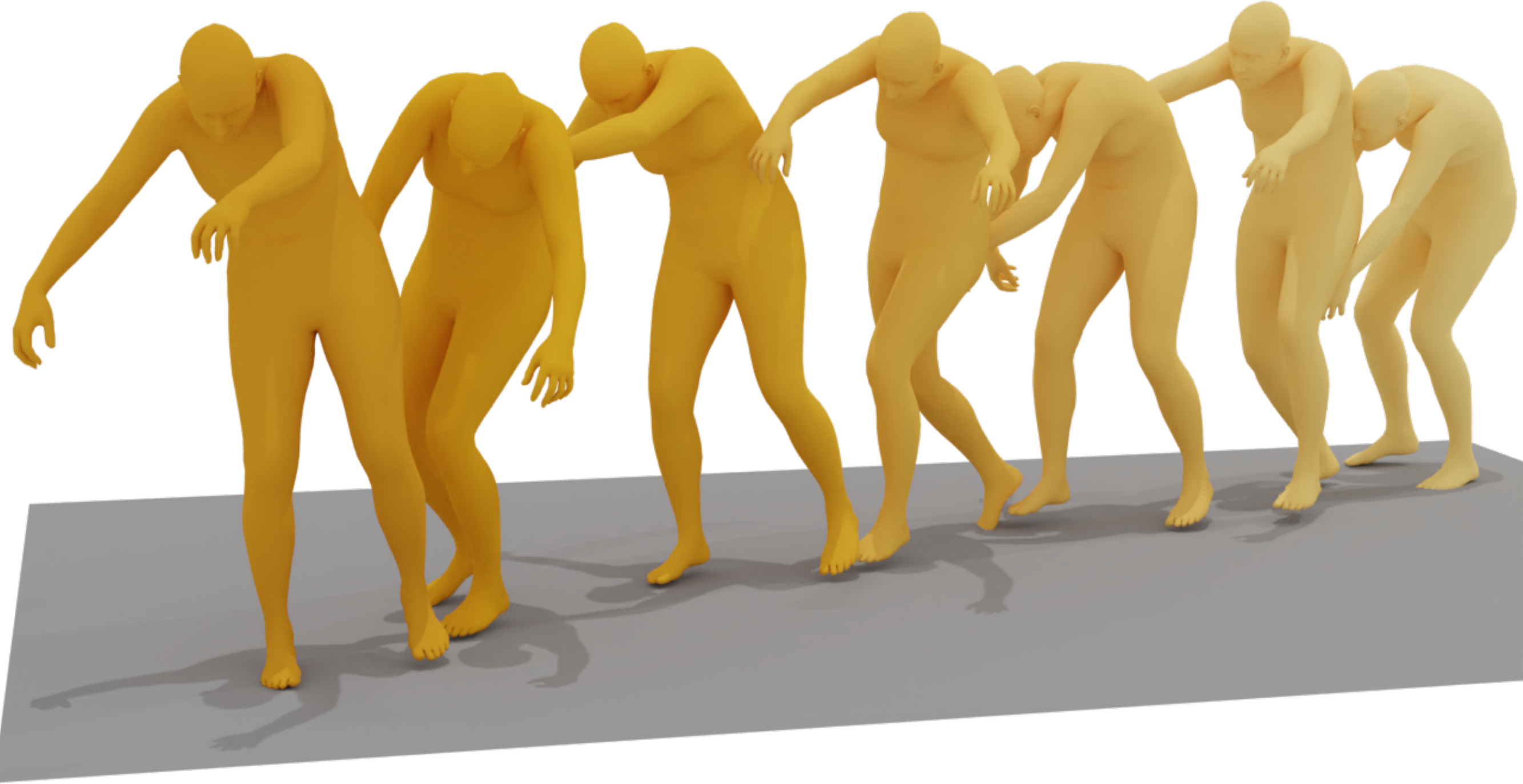}
        \end{minipage}
        \caption{A zombie slowly dragging its feet forward, arms outstretched, letting out a low groan.}
    \end{subfigure}
    \hfill
    \begin{subfigure}[b]{0.32\textwidth}
        \centering
        \begin{minipage}[t][4.2cm][t]{\textwidth}  
            \centering
            \includegraphics[width=\linewidth]{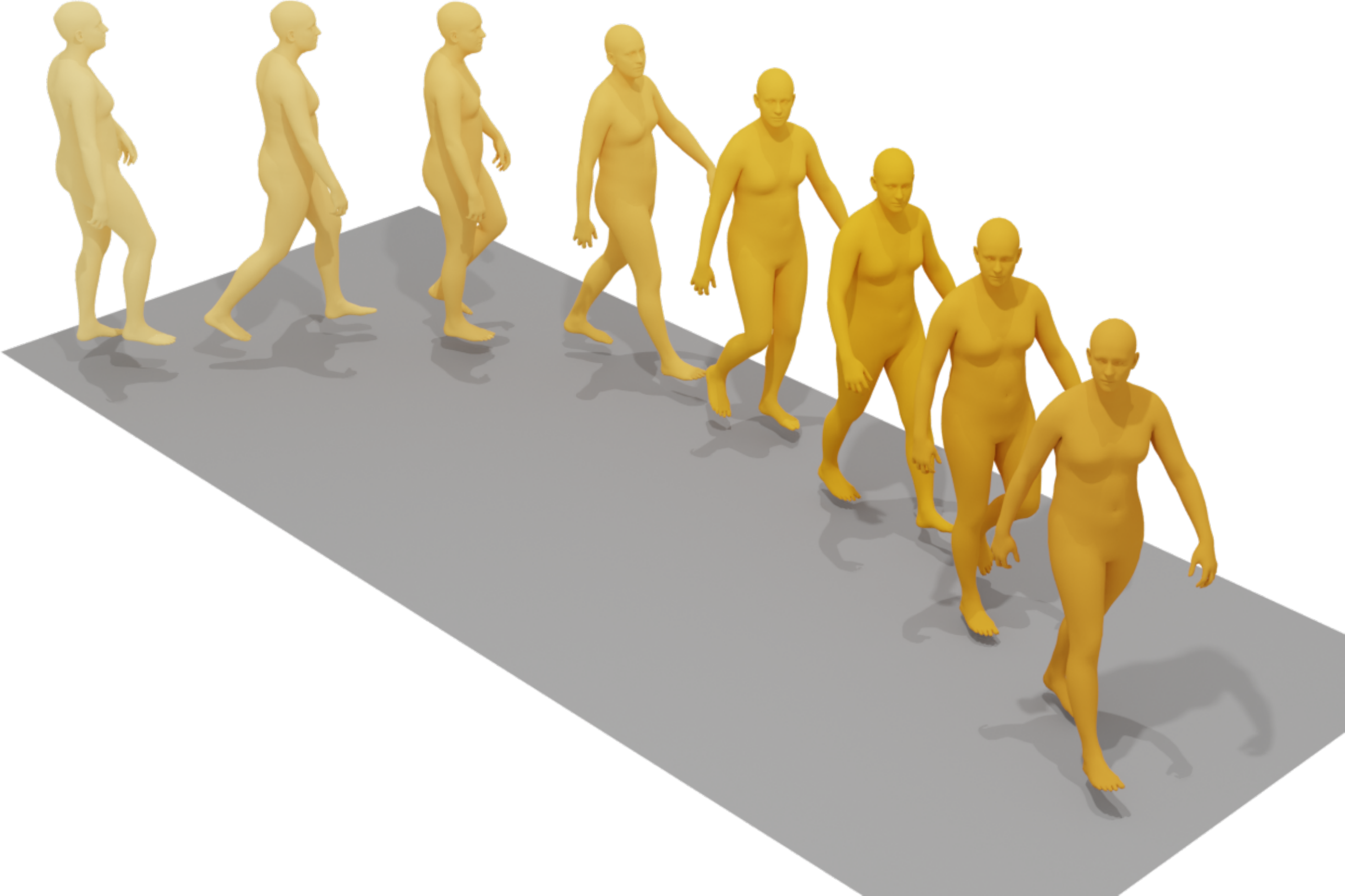}
        \end{minipage}
        \caption{An obese middle-aged male security guard, walking and looking around.}
    \end{subfigure}
    \hfill
    \begin{subfigure}[b]{0.32\textwidth}
        \centering
        \begin{minipage}[t][4.2cm][t]{\textwidth}  
            \centering
            \includegraphics[height=3.2cm]{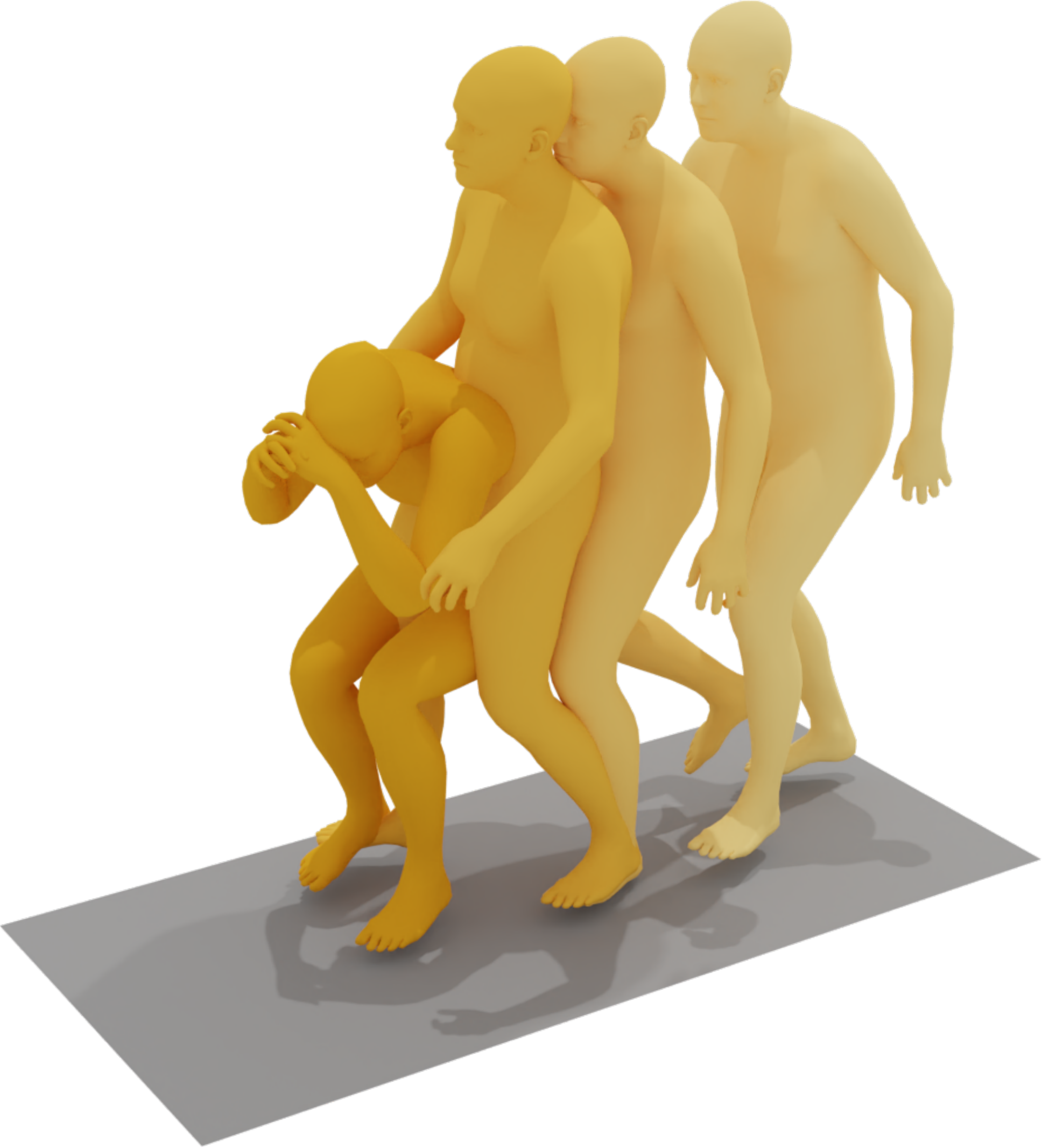}  
        \end{minipage}
        \caption{A man of average build who looked lost was walking along the street when a giant pie hit his head.}
    \end{subfigure}

    \caption{Zero-shot Text-to-Motion Generation Results on MotionMillion-Eval~\cite{fan2025go} prompts.}
    \vspace{-0.1cm}
    \label{fig:t2m_pic}
\end{figure*}

We show some examples of zero-shot text-to-motion on MotionMillion-Eval~\cite{fan2025go} prompts.
As shown in~\cref{fig:t2m_pic}, our model has robust performance in generating plausible and semantic aligned motions from unseen complex compositional textual descriptions.
We show more results and analysis in the appendix, where we also note some initial emergent model behavior, like motion generation with non-English language text input despite the model never having seen non-English text during unified training.

\subsection{Modality Specific Parameters Ablation}
\begin{table*}[t]
    \centering
    {
    \begin{tabular}{lccL{0.01cm}ccL{0.01cm}ccc}
        \toprule
        \multirow{2}{*}{Methods} & \multicolumn{2}{c}{Language-Only Performance} && \multicolumn{2}{c}{Motion-to-Text} && \multicolumn{2}{c}{Text-to-Motion} \\
        \cmidrule{2-3}\cmidrule{5-6}\cmidrule{8-9}
         & MMLU~\cite{hendrycks2020measuring}$\uparrow$ & IFEval~\cite{zhou2023instruction}$\uparrow$ && R@3$\uparrow$ & MM-D$\downarrow$ && FID$\downarrow$ & R@3$\uparrow$ \\
        \midrule
        Llama3.2-1B-Instruct & 49.3 & 59.5 && - & - && - & -\\ 
        Llama3.2-3B-Instruct & 63.4 & 77.4 && - & - && - & -\\
        \midrule
        Real Motion & - & - && 0.9866 & 0.7016 && - & 0.9866 \\
        \midrule
        \name-1B w/o MoT & 26.6 & 23.9 && 0.9380 & 0.7241 && 63.215 & 0.8110\\
        \name-3B w/o MoT & 24.9 & 22.3 && 0.9412 & 0.7148  && 46.174 & 0.8307\\
        \midrule
        \name-1B (our) & 49.3 & 59.5 && 0.9393 & 0.7136 && 27.361 & 0.9332\\
        \name-3B (our) & 63.4 & 77.4&& 0.9422& 0.7132& &19.893 & 0.9594\\
        \bottomrule
    \end{tabular}}
    \caption{\textbf{Ablation of Transformer Design Choice.} We evaluate our models on the test split ($\sim$ 30K samples) of our large-scale motion-text dataset. We follow the evaluator design in~\cite{guo2025snapmogen}, text-to-motion protocols in~\cite{guo2022generating}, and motion-to-text protocols in~\cite{guo2022tm2t}.}
    \label{tab:mot_abl}
    \vspace{-0.1cm}
\end{table*}
A key design principle of \name is to separate model parameters based on modality, \ie Mixture-of-Transformer.
While this design can readily preserve the language capability of the base LLMs by freezing the text module, it remains to be validated whether this approach benefits the multimodal learning process.
In this section, we conduct an ablation study by directly fine-tuning the full weights of the LLM instead of using MoT, which is mostly aligned with the design choices in previous works~\cite{jiang2023motiongpt,wang2024motiongpt,cao2025being}
To ensure reliable assessment, we train the evaluator following the protocol in~\cite{guo2025snapmogen} on our test split of the large-scale motion-text dataset.
As shown in~\cref{tab:mot_abl}, without any text-only corpus, ``\name~w/o~MoT'' suffers severe catastrophic forgetting, with MMLU~\cite{hendrycks2020measuring} and IFEval~\cite{zhou2023instruction} scores collapsing to near-random levels ($\leq$25~and~$\leq$30,~respectively), indicating an almost complete loss of basic world knowledge and instruction-following ability.
Furthermore, jointly optimizing flow-matching and discrete next-token prediction on the same parameters degrades training dynamics and motion generation, similar to observations in unified vision-language model training~\cite{deng2025emerging}.


\section{Limitations And Discussions}
Benefiting from the Mixture-of-Transformers design, the large model doubles the parameter count while keeping the per-token activation cost during inference identical to the base LLM. However, this architecture still substantially increases training cost compared with full-weight tuning under the same dataset setting. 
Although the continuous motion–token autoregressive formulation yields better results than discrete motion codecs, we find that its training dynamics require careful tuning. 
In future work, we plan to incorporate more instruction-tuning tasks, such as motion editing~\cite{athanasiou2024motionfix} and motion QA~\cite{li2025imore,endo2023motion}, to further leverage our large motion–language model and enable a broader range of downstream tasks within a single end-to-end framework.

\section{Conclusion}
In this paper, we introduce \name, the first large-scale motion-language model pretrained with a continuous autoregressive framework, enabling unified motion generation and understanding while preserving the base LLM's language capabilities. 
To achieve a seamless integration of motion and language, we design a modality-specific Mixture-of-Transformers architecture and freeze the text branch parameters to maintain linguistic proficiency.
Unlike previous approaches that discretize human motion, our method employs a continuous, causal VAE-based motion codec and utilizes flow-matching to model the next-token prediction distribution. 
Leveraging a large-scale motion-text dataset and the advanced language priors of base LLMs, our results show that \name establishes a strong foundation for next-generation motion-language models, bridging motion generation and understanding within a unified continuous autoregressive paradigm.

\section*{Acknowledgments}
This work is performed during the full-time and part-time internship at Meta in 2025. A part of this work was supported by NSF CAREER grant \#2143576.
{
    \small
    \bibliographystyle{ieeenat_fullname}
    \bibliography{main}

@String(CVPR= {IEEE Conf. Comput. Vis. Pattern Recog.})

@String(ICCV= {Int. Conf. Comput. Vis.})

@String(ECCV= {Eur. Conf. Comput. Vis.})

@String(AAAI = {AAAI})

@String(CVPR  = {CVPR})

@String(ICCV  = {ICCV})

@String(ECCV  = {ECCV})

@article{fan2025go,
  title={Go to zero: Towards zero-shot motion generation with million-scale data},
  author={Fan, Ke and Lu, Shunlin and Dai, Minyue and Yu, Runyi and Xiao, Lixing and Dou, Zhiyang and Dong, Junting and Ma, Lizhuang and Wang, Jingbo},
  journal={arXiv preprint arXiv:2507.07095},
  year={2025}
}

@article{xiao2025motionstreamer,
  title={MotionStreamer: Streaming Motion Generation via Diffusion-based Autoregressive Model in Causal Latent Space},
  author={Xiao, Lixing and Lu, Shunlin and Pi, Huaijin and Fan, Ke and Pan, Liang and Zhou, Yueer and Feng, Ziyong and Zhou, Xiaowei and Peng, Sida and Wang, Jingbo},
  journal={arXiv preprint arXiv:2503.15451},
  year={2025}
}

@article{hendrycks2020measuring,
  title={Measuring massive multitask language understanding},
  author={Hendrycks, Dan and Burns, Collin and Basart, Steven and Zou, Andy and Mazeika, Mantas and Song, Dawn and Steinhardt, Jacob},
  journal={arXiv preprint arXiv:2009.03300},
  year={2020}
}

@article{zhou2023instruction,
  title={Instruction-following evaluation for large language models},
  author={Zhou, Jeffrey and Lu, Tianjian and Mishra, Swaroop and Brahma, Siddhartha and Basu, Sujoy and Luan, Yi and Zhou, Denny and Hou, Le},
  journal={arXiv preprint arXiv:2311.07911},
  year={2023}
}

@inproceedings{endo2023motion,
  title={Motion question answering via modular motion programs},
  author={Endo, Mark and Hsu, Joy and Li, Jiaman and Wu, Jiajun},
  booktitle={International Conference on Machine Learning},
  pages={9312--9328},
  year={2023},
  organization={PMLR}
}

@article{li2025imore,
  title={IMoRe: Implicit Program-Guided Reasoning for Human Motion Q\&A},
  author={Li, Chen and Sugandhika, Chinthani and Ee, Yeo Keat and Peh, Eric and Zhang, Hao and Yang, Hong and Rajan, Deepu and Fernando, Basura},
  journal={arXiv preprint arXiv:2508.01984},
  year={2025}
}

@article{deng2025emerging,
  title={Emerging properties in unified multimodal pretraining},
  author={Deng, Chaorui and Zhu, Deyao and Li, Kunchang and Gou, Chenhui and Li, Feng and Wang, Zeyu and Zhong, Shu and Yu, Weihao and Nie, Xiaonan and Song, Ziang and others},
  journal={arXiv preprint arXiv:2505.14683},
  year={2025}
}

@article{team2025nextstep,
  title={NextStep-1: Toward autoregressive image generation with continuous tokens at scale},
  author={Team, NextStep and Han, Chunrui and Li, Guopeng and Wu, Jingwei and Sun, Quan and Cai, Yan and Peng, Yuang and Ge, Zheng and Zhou, Deyu and Tang, Haomiao and others},
  journal={arXiv preprint arXiv:2508.10711},
  year={2025}
}

@article{xie2025show,
  title={Show-o2: Improved Native Unified Multimodal Models},
  author={Xie, Jinheng and Yang, Zhenheng and Shou, Mike Zheng},
  journal={arXiv preprint arXiv:2506.15564},
  year={2025}
}

@article{xu2025qwen2,
  title={Qwen2. 5-omni technical report},
  author={Xu, Jin and Guo, Zhifang and He, Jinzheng and Hu, Hangrui and He, Ting and Bai, Shuai and Chen, Keqin and Wang, Jialin and Fan, Yang and Dang, Kai and others},
  journal={arXiv preprint arXiv:2503.20215},
  year={2025}
}

@misc{xu2025qwen3omni,
      title={Qwen3-Omni Technical Report}, 
      author={Jin Xu and Zhifang Guo and Hangrui Hu and Yunfei Chu and Xiong Wang and Jinzheng He and Yuxuan Wang and Xian Shi and Ting He and Xinfa Zhu and Yuanjun Lv and Yongqi Wang and Dake Guo and He Wang and Linhan Ma and Pei Zhang and Xinyu Zhang and Hongkun Hao and Zishan Guo and Baosong Yang and Bin Zhang and Ziyang Ma and Xipin Wei and Shuai Bai and Keqin Chen and Xuejing Liu and Peng Wang and Mingkun Yang and Dayiheng Liu and Xingzhang Ren and Bo Zheng and Rui Men and Fan Zhou and Bowen Yu and Jianxin Yang and Le Yu and Jingren Zhou and Junyang Lin},
      year={2025},
      eprint={2509.17765},
      archivePrefix={arXiv},
      primaryClass={cs.CL},
      url={https://arxiv.org/abs/2509.17765}, 
}

@InProceedings{cao2025being,
    author    = {Cao, Bin and Zheng, Sipeng and Wang, Ye and Xia, Lujie and Wei, Qianshan and Jin, Qin and Liu, Jing and Lu, Zongqing},
    title     = {MotionCtrl: A Real-time Controllable Vision-Language-Motion Model},
    booktitle = {Proceedings of the IEEE/CVF International Conference on Computer Vision (ICCV)},
    month     = {October},
    year      = {2025},
    pages     = {12253-12262}
}

@article{jiang2023motiongpt,
  title={Motiongpt: Human motion as a foreign language},
  author={Jiang, Biao and Chen, Xin and Liu, Wen and Yu, Jingyi and Yu, Gang and Chen, Tao},
  journal={Advances in Neural Information Processing Systems},
  volume={36},
  pages={20067--20079},
  year={2023}
}

@article{wang2024motiongpt,
  title={Motiongpt-2: A general-purpose motion-language model for motion generation and understanding},
  author={Wang, Yuan and Huang, Di and Zhang, Yaqi and Ouyang, Wanli and Jiao, Jile and Feng, Xuetao and Zhou, Yan and Wan, Pengfei and Tang, Shixiang and Xu, Dan},
  journal={arXiv preprint arXiv:2410.21747},
  year={2024}
}

@article{wu2024motion,
  title={Motion-agent: A conversational framework for human motion generation with llms},
  author={Wu, Qi and Zhao, Yubo and Wang, Yifan and Liu, Xinhang and Tai, Yu-Wing and Tang, Chi-Keung},
  journal={arXiv preprint arXiv:2405.17013},
  year={2024}
}

@inproceedings{jiang2024motionchain,
  title={Motionchain: Conversational motion controllers via multimodal prompts},
  author={Jiang, Biao and Chen, Xin and Zhang, Chi and Yin, Fukun and Li, Zhuoyuan and Yu, Gang and Fan, Jiayuan},
  booktitle={European Conference on Computer Vision},
  pages={54--74},
  year={2024},
  organization={Springer}
}

@inproceedings{guo2024momask,
  title={Momask: Generative masked modeling of 3d human motions},
  author={Guo, Chuan and Mu, Yuxuan and Javed, Muhammad Gohar and Wang, Sen and Cheng, Li},
  booktitle={Proceedings of the IEEE/CVF Conference on Computer Vision and Pattern Recognition},
  pages={1900--1910},
  year={2024}
}

@inproceedings{lu2025scamo,
  title={Scamo: Exploring the scaling law in autoregressive motion generation model},
  author={Lu, Shunlin and Wang, Jingbo and Lu, Zeyu and Chen, Ling-Hao and Dai, Wenxun and Dong, Junting and Dou, Zhiyang and Dai, Bo and Zhang, Ruimao},
  booktitle={Proceedings of the Computer Vision and Pattern Recognition Conference},
  pages={27872--27882},
  year={2025}
}

@article{li2024autoregressive,
  title={Autoregressive image generation without vector quantization},
  author={Li, Tianhong and Tian, Yonglong and Li, He and Deng, Mingyang and He, Kaiming},
  journal={Advances in Neural Information Processing Systems},
  volume={37},
  pages={56424--56445},
  year={2024}
}

@inproceedings{guo2022generating,
  title={Generating diverse and natural 3d human motions from text},
  author={Guo, Chuan and Zou, Shihao and Zuo, Xinxin and Wang, Sen and Ji, Wei and Li, Xingyu and Cheng, Li},
  booktitle={Proceedings of the IEEE/CVF conference on computer vision and pattern recognition},
  pages={5152--5161},
  year={2022}
}

@inproceedings{guo2022tm2t,
  title={Tm2t: Stochastic and tokenized modeling for the reciprocal generation of 3d human motions and texts},
  author={Guo, Chuan and Zuo, Xinxin and Wang, Sen and Cheng, Li},
  booktitle={European Conference on Computer Vision},
  pages={580--597},
  year={2022},
  organization={Springer}
}

@article{xie2025x,
  title={X-Streamer: Unified Human World Modeling with Audiovisual Interaction},
  author={Xie, You and Gu, Tianpei and Li, Zenan and Zhang, Chenxu and Song, Guoxian and Zhao, Xiaochen and Liang, Chao and Jiang, Jianwen and Xu, Hongyi and Luo, Linjie},
  journal={arXiv preprint arXiv:2509.21574},
  year={2025}
}

@inproceedings{zhou2019continuity,
  title={On the continuity of rotation representations in neural networks},
  author={Zhou, Yi and Barnes, Connelly and Lu, Jingwan and Yang, Jimei and Li, Hao},
  booktitle={Proceedings of the IEEE/CVF conference on computer vision and pattern recognition},
  pages={5745--5753},
  year={2019}
}

@article{brown2020language,
  title={Language models are few-shot learners},
  author={Brown, Tom and Mann, Benjamin and Ryder, Nick and Subbiah, Melanie and Kaplan, Jared D and Dhariwal, Prafulla and Neelakantan, Arvind and Shyam, Pranav and Sastry, Girish and Askell, Amanda and others},
  journal={Advances in neural information processing systems},
  volume={33},
  pages={1877--1901},
  year={2020}
}

@article{touvron2023llama,
  title={Llama: Open and efficient foundation language models},
  author={Touvron, Hugo and Lavril, Thibaut and Izacard, Gautier and Martinet, Xavier and Lachaux, Marie-Anne and Lacroix, Timoth{\'e}e and Rozi{\`e}re, Baptiste and Goyal, Naman and Hambro, Eric and Azhar, Faisal and others},
  journal={arXiv preprint arXiv:2302.13971},
  year={2023}
}

@article{team2024chameleon,
  title={Chameleon: Mixed-modal early-fusion foundation models},
  author={Team, Chameleon},
  journal={arXiv preprint arXiv:2405.09818},
  year={2024}
}

@article{tian2024mm,
  title={Mm-interleaved: Interleaved image-text generative modeling via multi-modal feature synchronizer},
  author={Tian, Changyao and Zhu, Xizhou and Xiong, Yuwen and Wang, Weiyun and Chen, Zhe and Wang, Wenhai and Chen, Yuntao and Lu, Lewei and Lu, Tong and Zhou, Jie and others},
  journal={arXiv preprint arXiv:2401.10208},
  year={2024}
}

@article{wang2024emu3,
  title={Emu3: Next-token prediction is all you need},
  author={Wang, Xinlong and Zhang, Xiaosong and Luo, Zhengxiong and Sun, Quan and Cui, Yufeng and Wang, Jinsheng and Zhang, Fan and Wang, Yueze and Li, Zhen and Yu, Qiying and others},
  journal={arXiv preprint arXiv:2409.18869},
  year={2024}
}

@inproceedings{wu2025janus,
  title={Janus: Decoupling visual encoding for unified multimodal understanding and generation},
  author={Wu, Chengyue and Chen, Xiaokang and Wu, Zhiyu and Ma, Yiyang and Liu, Xingchao and Pan, Zizheng and Liu, Wen and Xie, Zhenda and Yu, Xingkai and Ruan, Chong and others},
  booktitle={Proceedings of the Computer Vision and Pattern Recognition Conference},
  pages={12966--12977},
  year={2025}
}

@article{lu2022unified,
  title={Unified-io: A unified model for vision, language, and multi-modal tasks},
  author={Lu, Jiasen and Clark, Christopher and Zellers, Rowan and Mottaghi, Roozbeh and Kembhavi, Aniruddha},
  journal={arXiv preprint arXiv:2206.08916},
  year={2022}
}

@article{chen2025janus,
  title={Janus-pro: Unified multimodal understanding and generation with data and model scaling},
  author={Chen, Xiaokang and Wu, Zhiyu and Liu, Xingchao and Pan, Zizheng and Liu, Wen and Xie, Zhenda and Yu, Xingkai and Ruan, Chong},
  journal={arXiv preprint arXiv:2501.17811},
  year={2025}
}

@article{zhou2024transfusion,
  title={Transfusion: Predict the next token and diffuse images with one multi-modal model},
  author={Zhou, Chunting and Yu, Lili and Babu, Arun and Tirumala, Kushal and Yasunaga, Michihiro and Shamis, Leonid and Kahn, Jacob and Ma, Xuezhe and Zettlemoyer, Luke and Levy, Omer},
  journal={arXiv preprint arXiv:2408.11039},
  year={2024}
}

@article{xie2024show,
  title={Show-o: One single transformer to unify multimodal understanding and generation},
  author={Xie, Jinheng and Mao, Weijia and Bai, Zechen and Zhang, David Junhao and Wang, Weihao and Lin, Kevin Qinghong and Gu, Yuchao and Chen, Zhijie and Yang, Zhenheng and Shou, Mike Zheng},
  journal={arXiv preprint arXiv:2408.12528},
  year={2024}
}

@article{zhao2024monoformer,
  title={Monoformer: One transformer for both diffusion and autoregression},
  author={Zhao, Chuyang and Song, Yuxing and Wang, Wenhao and Feng, Haocheng and Ding, Errui and Sun, Yifan and Xiao, Xinyan and Wang, Jingdong},
  journal={arXiv preprint arXiv:2409.16280},
  year={2024}
}

@article{shi2024lmfusion,
  title={LMFusion: Adapting Pretrained Language Models for Multimodal Generation},
  author={Shi, Weijia and Han, Xiaochuang and Zhou, Chunting and Liang, Weixin and Lin, Xi Victoria and Zettlemoyer, Luke and Yu, Lili},
  journal={arXiv preprint arXiv:2412.15188},
  year={2024}
}

@article{liao2025mogao,
  title={Mogao: An omni foundation model for interleaved multi-modal generation},
  author={Liao, Chao and Liu, Liyang and Wang, Xun and Luo, Zhengxiong and Zhang, Xinyu and Zhao, Wenliang and Wu, Jie and Li, Liang and Tian, Zhi and Huang, Weilin},
  journal={arXiv preprint arXiv:2505.05472},
  year={2025}
}

@inproceedings{ma2025janusflow,
  title={Janusflow: Harmonizing autoregression and rectified flow for unified multimodal understanding and generation},
  author={Ma, Yiyang and Liu, Xingchao and Chen, Xiaokang and Liu, Wen and Wu, Chengyue and Wu, Zhiyu and Pan, Zizheng and Xie, Zhenda and Zhang, Haowei and Yu, Xingkai and others},
  booktitle={Proceedings of the Computer Vision and Pattern Recognition Conference},
  pages={7739--7751},
  year={2025}
}

@article{sun2024multimodal,
  title={Multimodal latent language modeling with next-token diffusion},
  author={Sun, Yutao and Bao, Hangbo and Wang, Wenhui and Peng, Zhiliang and Dong, Li and Huang, Shaohan and Wang, Jianyong and Wei, Furu},
  journal={arXiv preprint arXiv:2412.08635},
  year={2024}
}

@article{zhang2019root,
  title={Root mean square layer normalization},
  author={Zhang, Biao and Sennrich, Rico},
  journal={Advances in neural information processing systems},
  volume={32},
  year={2019}
}

@article{lipman2022flow,
  title={Flow matching for generative modeling},
  author={Lipman, Yaron and Chen, Ricky TQ and Ben-Hamu, Heli and Nickel, Maximilian and Le, Matt},
  journal={arXiv preprint arXiv:2210.02747},
  year={2022}
}

@article{liang2024intergen,
  title={Intergen: Diffusion-based multi-human motion generation under complex interactions},
  author={Liang, Han and Zhang, Wenqian and Li, Wenxuan and Yu, Jingyi and Xu, Lan},
  journal={International Journal of Computer Vision},
  volume={132},
  number={9},
  pages={3463--3483},
  year={2024},
  publisher={Springer}
}

@inproceedings{punnakkal2021babel,
  title={BABEL: Bodies, action and behavior with english labels},
  author={Punnakkal, Abhinanda R and Chandrasekaran, Arjun and Athanasiou, Nikos and Quiros-Ramirez, Alejandra and Black, Michael J},
  booktitle={Proceedings of the IEEE/CVF conference on computer vision and pattern recognition},
  pages={722--731},
  year={2021}
}

@article{lin2023motion,
  title={Motion-x: A large-scale 3d expressive whole-body human motion dataset},
  author={Lin, Jing and Zeng, Ailing and Lu, Shunlin and Cai, Yuanhao and Zhang, Ruimao and Wang, Haoqian and Zhang, Lei},
  journal={Advances in Neural Information Processing Systems},
  volume={36},
  pages={25268--25280},
  year={2023}
}

@inproceedings{li2023finedance,
  title={Finedance: A fine-grained choreography dataset for 3d full body dance generation},
  author={Li, Ronghui and Zhao, Junfan and Zhang, Yachao and Su, Mingyang and Ren, Zeping and Zhang, Han and Tang, Yansong and Li, Xiu},
  booktitle={Proceedings of the IEEE/CVF International Conference on Computer Vision},
  pages={10234--10243},
  year={2023}
}

@inproceedings{yin2023hi4d,
  title={Hi4d: 4d instance segmentation of close human interaction},
  author={Yin, Yifei and Guo, Chen and Kaufmann, Manuel and Zarate, Juan Jose and Song, Jie and Hilliges, Otmar},
  booktitle={Proceedings of the IEEE/CVF Conference on Computer Vision and Pattern Recognition},
  pages={17016--17027},
  year={2023}
}

@inproceedings{fieraru2021learning,
  title={Learning complex 3D human self-contact},
  author={Fieraru, Mihai and Zanfir, Mihai and Oneata, Elisabeta and Popa, Alin-Ionut and Olaru, Vlad and Sminchisescu, Cristian},
  booktitle={Proceedings of the AAAI Conference on Artificial Intelligence},
  volume={35},
  number={2},
  pages={1343--1351},
  year={2021}
}

@article{mclean2025embody,
  title={Embody 3D: A Large-scale Multimodal Motion and Behavior Dataset},
  author={McLean, Claire and Meendering, Makenzie and Swartz, Tristan and Gabbay, Orri and Olsen, Alexandra and Jacobs, Rachel and Rosen, Nicholas and de Bree, Philippe and Garcia, Tony and Merrill, Gadsden and others},
  journal={arXiv preprint arXiv:2510.16258},
  year={2025}
}

@article{tevet2022human,
  title={Human motion diffusion model},
  author={Tevet, Guy and Raab, Sigal and Gordon, Brian and Shafir, Yonatan and Cohen-Or, Daniel and Bermano, Amit H},
  journal={arXiv preprint arXiv:2209.14916},
  year={2022}
}

@inproceedings{chen2023executing,
  title={Executing your commands via motion diffusion in latent space},
  author={Chen, Xin and Jiang, Biao and Liu, Wen and Huang, Zilong and Fu, Bin and Chen, Tao and Yu, Gang},
  booktitle={Proceedings of the IEEE/CVF conference on computer vision and pattern recognition},
  pages={18000--18010},
  year={2023}
}

@inproceedings{zhang2023generating,
  title={Generating human motion from textual descriptions with discrete representations},
  author={Zhang, Jianrong and Zhang, Yangsong and Cun, Xiaodong and Zhang, Yong and Zhao, Hongwei and Lu, Hongtao and Shen, Xi and Shan, Ying},
  booktitle={Proceedings of the IEEE/CVF conference on computer vision and pattern recognition},
  pages={14730--14740},
  year={2023}
}

@inproceedings{zhong2023attt2m,
  title={Attt2m: Text-driven human motion generation with multi-perspective attention mechanism},
  author={Zhong, Chongyang and Hu, Lei and Zhang, Zihao and Xia, Shihong},
  booktitle={Proceedings of the IEEE/CVF international conference on computer vision},
  pages={509--519},
  year={2023}
}

@inproceedings{chuan2022tm2t,
  title={TM2T: Stochastic and Tokenized Modeling for the Reciprocal Generation of 3D Human Motions and Texts},
  author={Guo, Chuan and Zuo, Xinxin and Wang, Sen and Cheng, Li},
  booktitle={ECCV},
  year={2022}
}

@article{li2024lamp,
  title={Lamp: Language-motion pretraining for motion generation, retrieval, and captioning},
  author={Li, Zhe and Yuan, Weihao and He, Yisheng and Qiu, Lingteng and Zhu, Shenhao and Gu, Xiaodong and Shen, Weichao and Dong, Yuan and Dong, Zilong and Yang, Laurence T},
  journal={arXiv preprint arXiv:2410.07093},
  year={2024}
}

@article{wu2024mote,
  title={MoTe: Learning Motion-Text Diffusion Model for Multiple Generation Tasks},
  author={Wu, Yiming and Ji, Wei and Zheng, Kecheng and Wang, Zicheng and Xu, Dong},
  journal={arXiv preprint arXiv:2411.19786},
  year={2024}
}

@article{vaswani2017attention,
  title={Attention is all you need},
  author={Vaswani, Ashish and Shazeer, Noam and Parmar, Niki and Uszkoreit, Jakob and Jones, Llion and Gomez, Aidan N and Kaiser, {\L}ukasz and Polosukhin, Illia},
  journal={Advances in neural information processing systems},
  volume={30},
  year={2017}
}

@InProceedings{pmlr-v202-li23q,
  title = 	 {{BLIP}-2: Bootstrapping Language-Image Pre-training with Frozen Image Encoders and Large Language Models},
  author =       {Li, Junnan and Li, Dongxu and Savarese, Silvio and Hoi, Steven},
  booktitle = 	 {Proceedings of the 40th International Conference on Machine Learning},
  pages = 	 {19730--19742},
  year = 	 {2023},
  editor = 	 {Krause, Andreas and Brunskill, Emma and Cho, Kyunghyun and Engelhardt, Barbara and Sabato, Sivan and Scarlett, Jonathan},
  volume = 	 {202},
  series = 	 {Proceedings of Machine Learning Research},
  month = 	 {23--29 Jul},
  publisher =    {PMLR},
  url = 	 {https://proceedings.mlr.press/v202/li23q.html}
}

@article{liu2023visual,
  title={Visual instruction tuning},
  author={Liu, Haotian and Li, Chunyuan and Wu, Qingyang and Lee, Yong Jae},
  journal={Advances in neural information processing systems},
  volume={36},
  pages={34892--34916},
  year={2023}
}

@inproceedings{ramesh2021zero,
  title={Zero-shot text-to-image generation},
  author={Ramesh, Aditya and Pavlov, Mikhail and Goh, Gabriel and Gray, Scott and Voss, Chelsea and Radford, Alec and Chen, Mark and Sutskever, Ilya},
  booktitle={International conference on machine learning},
  pages={8821--8831},
  year={2021},
  organization={Pmlr}
}

@inproceedings{shen2024world,
  title={World-grounded human motion recovery via gravity-view coordinates},
  author={Shen, Zehong and Pi, Huaijin and Xia, Yan and Cen, Zhi and Peng, Sida and Hu, Zechen and Bao, Hujun and Hu, Ruizhen and Zhou, Xiaowei},
  booktitle={SIGGRAPH Asia 2024 Conference Papers},
  pages={1--11},
  year={2024}
}

@inproceedings{li2019neural,
  title={Neural speech synthesis with transformer network},
  author={Li, Naihan and Liu, Shujie and Liu, Yanqing and Zhao, Sheng and Liu, Ming},
  booktitle={Proceedings of the AAAI conference on artificial intelligence},
  volume={33},
  number={01},
  pages={6706--6713},
  year={2019}
}

@inproceedings{ao2022speecht5,
  title={Speecht5: Unified-modal encoder-decoder pre-training for spoken language processing},
  author={Ao, Junyi and Wang, Rui and Zhou, Long and Wang, Chengyi and Ren, Shuo and Wu, Yu and Liu, Shujie and Ko, Tom and Li, Qing and Zhang, Yu and others},
  booktitle={Proceedings of the 60th annual meeting of the association for computational linguistics (volume 1: Long papers)},
  pages={5723--5738},
  year={2022}
}

@inproceedings{cho2025discord,
  title={DisCoRD: Discrete Tokens to Continuous Motion via Rectified Flow Decoding},
  author={Cho, Jungbin and Kim, Junwan and Kim, Jisoo and Kim, Minseo and Kang, Mingu and Hong, Sungeun and Oh, Tae-Hyun and Yu, Youngjae},
  booktitle={Proceedings of the IEEE/CVF International Conference on Computer Vision},
  pages={14602--14612},
  year={2025}
}

@article{zhu2025motiongpt3,
  title={MotionGPT3: Human Motion as a Second Modality},
  author={Zhu, Bingfan and Jiang, Biao and Wang, Sunyi and Tang, Shixiang and Chen, Tao and Luo, Linjie and Zheng, Youyi and Chen, Xin},
  journal={arXiv preprint arXiv:2506.24086},
  year={2025}
}

@inproceedings{vedantam2015cider,
  title={Cider: Consensus-based image description evaluation},
  author={Vedantam, Ramakrishna and Lawrence Zitnick, C and Parikh, Devi},
  booktitle={Proceedings of the IEEE conference on computer vision and pattern recognition},
  pages={4566--4575},
  year={2015}
}

@inproceedings{lin2004rouge,
  title={Rouge: A package for automatic evaluation of summaries},
  author={Lin, Chin-Yew},
  booktitle={Text summarization branches out},
  pages={74--81},
  year={2004}
}

@article{zhang2019bertscore,
  title={Bertscore: Evaluating text generation with bert},
  author={Zhang, Tianyi and Kishore, Varsha and Wu, Felix and Weinberger, Kilian Q and Artzi, Yoav},
  journal={arXiv preprint arXiv:1904.09675},
  year={2019}
}

@inproceedings{papineni2002bleu,
  title={Bleu: a method for automatic evaluation of machine translation},
  author={Papineni, Kishore and Roukos, Salim and Ward, Todd and Zhu, Wei-Jing},
  booktitle={Proceedings of the 40th annual meeting of the Association for Computational Linguistics},
  pages={311--318},
  year={2002}
}

@inproceedings{zhang2023remodiffuse,
  title={Remodiffuse: Retrieval-augmented motion diffusion model},
  author={Zhang, Mingyuan and Guo, Xinying and Pan, Liang and Cai, Zhongang and Hong, Fangzhou and Li, Huirong and Yang, Lei and Liu, Ziwei},
  booktitle={Proceedings of the IEEE/CVF International Conference on Computer Vision},
  pages={364--373},
  year={2023}
}

@article{cui2025emu3,
  title={Emu3. 5: Native Multimodal Models are World Learners},
  author={Cui, Yufeng and Chen, Honghao and Deng, Haoge and Huang, Xu and Li, Xinghang and Liu, Jirong and Liu, Yang and Luo, Zhuoyan and Wang, Jinsheng and Wang, Wenxuan and others},
  journal={arXiv preprint arXiv:2510.26583},
  year={2025}
}

@article{tan2025omni,
  title={Omni-video: Democratizing unified video understanding and generation},
  author={Tan, Zhiyu and Yang, Hao and Qin, Luozheng and Gong, Jia and Yang, Mengping and Li, Hao},
  journal={arXiv preprint arXiv:2507.06119},
  year={2025}
}

@article{wei2025univideo,
  title={UniVideo: Unified Understanding, Generation, and Editing for Videos},
  author={Wei, Cong and Liu, Quande and Ye, Zixuan and Wang, Qiulin and Wang, Xintao and Wan, Pengfei and Gai, Kun and Chen, Wenhu},
  journal={arXiv preprint arXiv:2510.08377},
  year={2025}
}

@article{xu2025mospa,
  title={Mospa: Human motion generation driven by spatial audio},
  author={Xu, Shuyang and Dou, Zhiyang and Shi, Mingyi and Pan, Liang and Ho, Leo and Wang, Jingbo and Liu, Yuan and Lin, Cheng and Ma, Yuexin and Wang, Wenping and others},
  journal={arXiv preprint arXiv:2507.11949},
  year={2025}
}

@article{pasini2024continuous,
  title={Continuous autoregressive models with noise augmentation avoid error accumulation},
  author={Pasini, Marco and Nistal, Javier and Lattner, Stefan and Fazekas, George},
  journal={arXiv preprint arXiv:2411.18447},
  year={2024}
}

@inproceedings{jiang2024scaling,
  title={Scaling up dynamic human-scene interaction modeling},
  author={Jiang, Nan and Zhang, Zhiyuan and Li, Hongjie and Ma, Xiaoxuan and Wang, Zan and Chen, Yixin and Liu, Tengyu and Zhu, Yixin and Huang, Siyuan},
  booktitle={Proceedings of the IEEE/CVF Conference on Computer Vision and Pattern Recognition},
  pages={1737--1747},
  year={2024}
}

@inproceedings{xu2024inter,
  title={Inter-x: Towards versatile human-human interaction analysis},
  author={Xu, Liang and Lv, Xintao and Yan, Yichao and Jin, Xin and Wu, Shuwen and Xu, Congsheng and Liu, Yifan and Zhou, Yizhou and Rao, Fengyun and Sheng, Xingdong and others},
  booktitle={Proceedings of the IEEE/CVF conference on computer vision and pattern recognition},
  pages={22260--22271},
  year={2024}
}

@InProceedings{Meng_2025_CVPR,
    author    = {Meng, Zichong and Xie, Yiming and Peng, Xiaogang and Han, Zeyu and Jiang, Huaizu},
    title     = {Rethinking Diffusion for Text-Driven Human Motion Generation: Redundant Representations, Evaluation, and Masked Autoregression},
    booktitle = {Proceedings of the IEEE/CVF Conference on Computer Vision and Pattern Recognition (CVPR)},
    month     = {June},
    year      = {2025},
    pages     = {27859-27871}
}

@article{hu2025hmvlm,
  title={HMVLM: Human Motion-Vision-Lanuage Model via MoE LoRA},
  author={Hu, Lei and Ye, Yongjing and Xia, Shihong},
  journal={arXiv preprint arXiv:2511.01463},
  year={2025}
}

@article{ke2025hyperspherical,
  title={Hyperspherical latents improve continuous-token autoregressive generation},
  author={Ke, Guolin and Xue, Hui},
  journal={arXiv preprint arXiv:2509.24335},
  year={2025}
}

@article{shao2025continuous,
  title={Continuous Autoregressive Language Models},
  author={Shao, Chenze and Li, Darren and Meng, Fandong and Zhou, Jie},
  journal={arXiv preprint arXiv:2510.27688},
  year={2025}
}

@article{guo2025deepseek,
  title={Deepseek-r1: Incentivizing reasoning capability in llms via reinforcement learning},
  author={Guo, Daya and Yang, Dejian and Zhang, Haowei and Song, Junxiao and Zhang, Ruoyu and Xu, Runxin and Zhu, Qihao and Ma, Shirong and Wang, Peiyi and Bi, Xiao and others},
  journal={arXiv preprint arXiv:2501.12948},
  year={2025}
}

@article{yue2025does,
  title={Does reinforcement learning really incentivize reasoning capacity in llms beyond the base model?},
  author={Yue, Yang and Chen, Zhiqi and Lu, Rui and Zhao, Andrew and Wang, Zhaokai and Song, Shiji and Huang, Gao},
  journal={arXiv preprint arXiv:2504.13837},
  year={2025}
}

@article{wang2025unirl,
  title={UniRL-Zero: Reinforcement Learning on Unified Models with Joint Language Model and Diffusion Model Experts},
  author={Wang, Fu-Yun and Zhang, Han and Gharbi, Michael and Li, Hongsheng and Park, Taesung},
  journal={arXiv preprint arXiv:2510.17937},
  year={2025}
}

@article{comanici2025gemini,
  title={Gemini 2.5: Pushing the frontier with advanced reasoning, multimodality, long context, and next generation agentic capabilities},
  author={Comanici, Gheorghe and Bieber, Eric and Schaekermann, Mike and Pasupat, Ice and Sachdeva, Noveen and Dhillon, Inderjit and Blistein, Marcel and Ram, Ori and Zhang, Dan and Rosen, Evan and others},
  journal={arXiv preprint arXiv:2507.06261},
  year={2025}
}

@article{ouyang2025motion,
  title={Motion-R1: Chain-of-Thought Reasoning and Reinforcement Learning for Human Motion Generation},
  author={Ouyang, Runqi and Li, Haoyun and Zhang, Zhenyuan and Wang, Xiaofeng and Zhu, Zheng and Huang, Guan and Wang, Xingang},
  journal={arXiv preprint arXiv:2506.10353},
  year={2025}
}

@article{zhang2025social,
  title={Social Agent: Mastering Dyadic Nonverbal Behavior Generation via Conversational LLM Agents},
  author={Zhang, Zeyi and Zhou, Yanju and Yao, Heyuan and Ao, Tenglong and Zhan, Xiaohang and Liu, Libin},
  journal={arXiv preprint arXiv:2510.04637},
  year={2025}
}

@inproceedings{jiang2025solami,
  title={Solami: Social vision-language-action modeling for immersive interaction with 3d autonomous characters},
  author={Jiang, Jianping and Xiao, Weiye and Lin, Zhengyu and Zhang, Huaizhong and Ren, Tianxiang and Gao, Yang and Lin, Zhiqian and Cai, Zhongang and Yang, Lei and Liu, Ziwei},
  booktitle={Proceedings of the Computer Vision and Pattern Recognition Conference},
  pages={26887--26898},
  year={2025}
}

@article{zhao2025navigating,
  title={Navigating Motion Agents in Dynamic and Cluttered Environments through LLM Reasoning},
  author={Zhao, Yubo and Wu, Qi and Wang, Yifan and Tai, Yu-Wing and Tang, Chi-Keung},
  journal={arXiv preprint arXiv:2503.07323},
  year={2025}
}

@InProceedings{Wang_2025_ICCV,
    author    = {Wang, Runqi and Ma, Caoyuan and Li, Guopeng and Xu, Hanrui and Li, Yuke and Wang, Zheng},
    title     = {You Think, You ACT: The New Task of Arbitrary Text to Motion Generation},
    booktitle = {Proceedings of the IEEE/CVF International Conference on Computer Vision (ICCV)},
    month     = {October},
    year      = {2025},
    pages     = {12012-12022}
}

@InProceedings{Fang_2025_CVPR,
    author    = {Fang, Qihang and Tang, Chengcheng and Tekin, Bugra and Ma, Shugao and Yang, Yanchao},
    title     = {HuMoCon: Concept Discovery for Human Motion Understanding},
    booktitle = {Proceedings of the IEEE/CVF Conference on Computer Vision and Pattern Recognition (CVPR)},
    month     = {June},
    year      = {2025},
    pages     = {7179-7190}
}

@article{guo2025snapmogen,
  title={SnapMoGen: Human Motion Generation from Expressive Texts},
  author={Guo, Chuan and Hwang, Inwoo and Wang, Jian and Zhou, Bing},
  journal={arXiv preprint arXiv:2507.09122},
  year={2025}
}

@inproceedings{athanasiou2024motionfix,
  title = {{MotionFix}: Text-Driven 3D Human Motion Editing},
  author = {Athanasiou, Nikos and Ceske, Alp{\'a}r and Diomataris, Markos and Black, Michael J. and Varol, G{\"u}l},
  booktitle = {SIGGRAPH Asia 2024 Conference Papers},
  year = {2024}
}

@article{liu2022flow,
  title={Flow straight and fast: Learning to generate and transfer data with rectified flow},
  author={Liu, Xingchao and Gong, Chengyue and Liu, Qiang},
  journal={arXiv preprint arXiv:2209.03003},
  year={2022}
}

@article{loshchilov2017decoupled,
  title={Decoupled weight decay regularization},
  author={Loshchilov, Ilya and Hutter, Frank},
  journal={arXiv preprint arXiv:1711.05101},
  year={2017}
}

@article{hendrycks2016gaussian,
  title={Gaussian Error Linear Units (Gelus)},
  author={Hendrycks, D},
  journal={arXiv preprint arXiv:1606.08415},
  year={2016}
}

@article{ramachandran2017searching,
  title={Searching for activation functions},
  author={Ramachandran, Prajit and Zoph, Barret and Le, Quoc V},
  journal={arXiv preprint arXiv:1710.05941},
  year={2017}
}

@article{mason2022real,
  title={Real-time style modelling of human locomotion via feature-wise transformations and local motion phases},
  author={Mason, Ian and Starke, Sebastian and Komura, Taku},
  journal={Proceedings of the ACM on Computer Graphics and Interactive Techniques},
  volume={5},
  number={1},
  pages={1--18},
  year={2022},
  publisher={ACM New York, NY, USA}
}

@misc{CombatMotion,
  title={AnimationGPT:An AIGC tool for generating game combat motion assets},
  author={Liao, Yihao and Fu, Yiyu and Cheng, Ziming and Wang, Jiangfeiyang},
  year={2024},
  howpublished={\url{https://github.com/fyyakaxyy/AnimationGPT}}
}

@article{zhang2025egoreact,
  title={EgoReAct: Egocentric Video-Driven 3D Human Reaction Generation},
  author={Zhang, Libo and Li, Zekun and Li, Tianyu and Cao, Zeyu and Xu, Rui and Long, Xiaoxiao and Wang, Wenjia and Wang, Jingbo and Liu, Yuan and Wang, Wenping and others},
  journal={arXiv preprint arXiv:2512.22808},
  year={2025}
}
}

\clearpage
\setcounter{page}{1}
\maketitlesupplementary

\section{Implementation Details}
In this section, we report all the details of \name's implementation, to support reproducibility. We further trained an 8B model and discuss all our model sizes - 1B, 3B and 8B. 

\subsection{Motion VAE}
We adopt the causal VAE architecture and training losses from MotionStreamer~\cite{xiao2025motionstreamer}.
The first 1K training iterations use a linear warmup learning rate schedule from 0 to 5e-5, followed by 3M iterations with a cosine decay schedule from 5e-5 to 0.
We use the AdamW optimizer~\cite{loshchilov2017decoupled} with [$\beta_1$, $\beta_2$] = [0.9, 0.95] and a batch size of 256. 
All VAE training runs use 8 A100 GPUs. 
For robustness in modeling continuous autoregressive tokens, we sample the variance for each VAE latent from $\mathbf{U}(0,\mathcal{C}_\sigma)$ where $\mathcal{C}_\sigma=0.01$, instead of predicting it from the network.

\subsection{Unified Motion-Language Model}
\paragraph{Mixture-of-Transformer.}
We used Llama3.2-1/3B-Instruct and Llama3.1-8B-Instruct as the base language model to build \name-1B/3B and 8B, respectively.
During training, all language-related parameters are frozen, except for the text embeddings of [BOM] and [EOM].
These special text token embeddings are initialized from the mean of the language codebook. .
The motion branch transformer parameters are initialized from the text branch transformer.
The motion adapter $\mathcal{P}(\cdot)$ is a two-layer MLP with GELU~\cite{hendrycks2016gaussian} activation and post-RMSNorm~\cite{zhang2019root}.

\paragraph{Flow Matching Head.}
We use the MLP head architecture design from MAR~\cite{li2024autoregressive}, with a hidden dimension of 1536 and 12 layers.
Before the output vectors of the Transformer serve as the condition for flow matching, we apply a two-layer MLP with GELU~\cite{hendrycks2016gaussian} activation and post-RMSNorm~\cite{zhang2019root} as the motion projector.
During inference, we use an ODE solver based on Euler integrator with 50 steps.

\begin{figure}[!ht]
    \centering
    \vspace{-0.5cm}
    \includegraphics[width=1\linewidth]{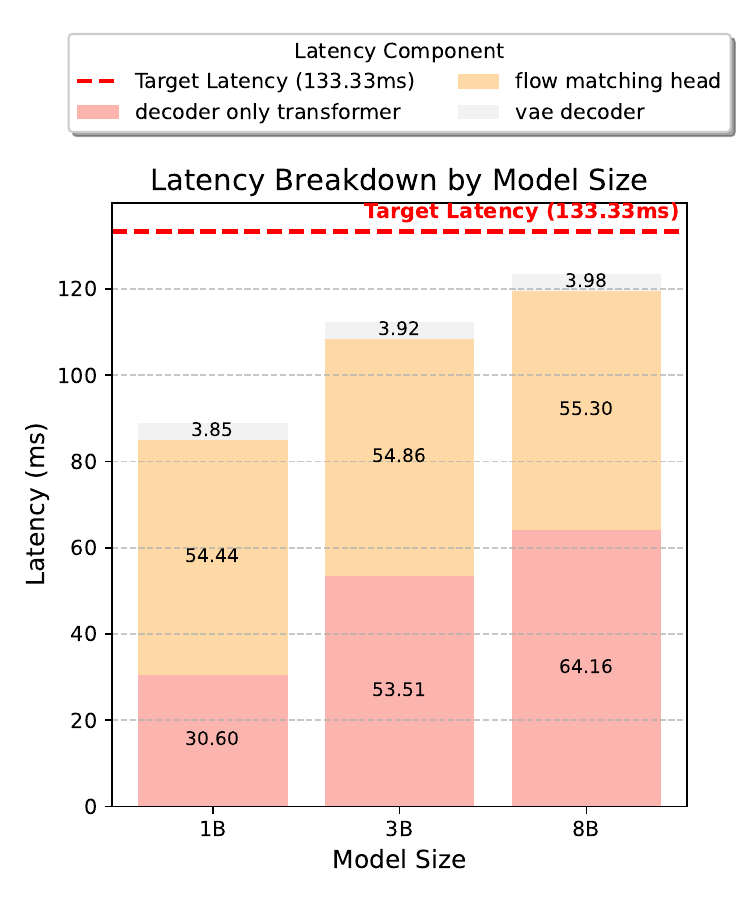}
    \vspace{-1cm}
    \caption{\textbf{Token Latency breakdown of Inference.} We compared the inference speed based on different model sizes. With  infrastructural optimizations, even 8B model can achieve real-time streaming motion generation. \textbf{Our VAE does 4x temporal downsampling. So the 7.5FPS token generation speed equal to 30FPS motion generation speed.}}
    \label{fig:speed}
\end{figure}
\paragraph{Motion Generation Exit Head.}
We follow TransformerTTS~\cite{li2019neural} and SpeechT5~\cite{ao2022speecht5}in using a simple MLP to predict the stop generation signal based on the decoder-only transformer output.
The MLP is structured by 5 Linear layers with Swish activation~\cite{ramachandran2017searching}.
We adopt a binary classification loss for stop prediction. 

\paragraph{Efficient Training and Inference}
To achieve more efficient training, we utilize DeepSpeed-Zero2~\footnote{\url{https://www.deepspeed.ai/tutorials/zero/}} to reduce the redundancy in optimizer states and updating gradients.
All the training is under BF16 precision via the AdamW optimizer~\cite{loshchilov2017decoupled} with [$\beta_1$, $\beta_2$] = [0.9, 0.95] and a batch size of 128.
The \name-1B and \name-3B are trained on 8 A100s.
The \name-8B are trained on 16 A100s.
To achieve real-time streaming motion generation on a single A100, we adopt serval infrastructural optimizations with few engineer efforts.
Specifically, we use KV-cache to reduce the computation of large decoder-only transformer and compile the cuda graph of flow matching sampling loop to remove kernel launch overhead.
We profile the cost of inference computation with batch size 4 in \cref{fig:speed} and, as shown, all the models with different sizes can achieve real-time streaming motion generation.
Since our motion causal VAE encodes human motion using a 4$\times$~temporal downsampling factor, the target token latency required to achieve real-time streaming motion generation is $1000 / 30 \times 4 = 133.33 \text{ ms}$. 

\begin{table}[!ht]
    \centering
    \resizebox{1\linewidth}{!}{
    \begin{tabular}{lccL{0.01cm}ccc}
        \toprule
        \multirow{2}{*}{Methods} & \multicolumn{2}{c}{Motion-to-Text} && \multicolumn{2}{c}{Text-to-Motion} \\
        \cmidrule{2-3}\cmidrule{5-6}
         & R@3$\uparrow$ & MM-D$\downarrow$ && FID$\downarrow$ & R@3$\uparrow$ \\
        \midrule
        Real Motion & 0.9866 & 0.7016 && - & 0.9866 \\
        \midrule
        \name-1B (our) &  0.9393 & 0.7136 && 27.361 & 0.9332\\
        \name-3B (our) &0.9422& 0.7132& &19.893 & 0.9594\\
        \name-8B (our) &0.9613& 0.7126& &18.935 & 0.9603\\
        \midrule
        \rowcolor{gray!15}\multicolumn{7}{l}{\makebox[\dimexpr\linewidth][l]{\textbf{Ablations based on \name-3B}}} \\
        use~\cite{xiao2025motionstreamer}VAE &  0.9221 & 0.7310 && 34.002 & 0.8936 \\
        only Stage1 & 0.9108 & 0.7443 && 80.912 & 0.7615\\
        only Stage1\&2 &0.9422& 0.7132& &22.524 &0.9521 \\
        \bottomrule
    \end{tabular}}
    \caption{\textbf{Ablation of Design Choice.} We evaluate our models on the test split ($\sim$ 30K samples) of our large-scale motion-text dataset. We follow the evaluator design in~\cite{guo2025snapmogen}, text-to-motion protocols in~\cite{guo2022generating}, and motion-to-text protocols in~\cite{guo2022tm2t}.
    We show that a) using a VAE with predicted variance significantly hurts motion generation, and b) our multistage training recipe progressively improves the model performance.
    } 
    \label{tab:abl_supp}
\end{table}
\section{More Ablation Studies}
In this section, we demonstrate more results and analysis to validate the effectiveness of \name design choices. 

\paragraph{Further Scale Up Model Size}
To explore the scalability of this solution, we design a 8B-version \name based on Llama-3.1-8B-Instruct.
Consistent with the findings in MotionMillion~\cite{fan2025go}, as shown in~\cref{tab:abl_supp}, we observe that scaling the model from 3B to 8B yields negligible improvements in FID and R-precision compared to the transition from 1B to 3B.

\paragraph{Traditional VAE \vs Our VAE}
Although prior works~\cite{team2025nextstep,sun2024multimodal,ke2025hyperspherical,shao2025continuous} have demonstrated that a robust VAE is crucial for continuous autoregressive generation, it remains unclear whether this conclusion generalizes to the motion modality.
Therefore, we further evaluate the validity and generalization of this observation in our motion-language setting in~\cref{tab:abl_supp}. 
Instead of sampling the variance from predefined distribution, we use the classic network-predicted variance VAE as in~\cite{xiao2025motionstreamer}.
The significant degradation of motion synthesis performance confirms that adding noise to ensure a robust VAE is essential for the continuous autoregressive paradigm.
However, we note that motion understanding performance is not affected by the VAE robustness.

\paragraph{Multi-Stage Training Recipe.}
We further evaluate the effect of our multi-stage training strategy, with results summarized in~\cref{tab:abl_supp} for the 3B model setting.
Across stages, we observe steady improvements in both motion fidelity and text–motion consistency, indicating that the staged optimization procedure effectively stabilizes the learning dynamics of large models.
By decomposing the training process into progressively more specialized phases, our approach mitigates early training instability, facilitates more reliable modality alignment, and ultimately leads to better overall zero-shot performance.

\section{Data Curation Details}
\paragraph{Annotation Prompt.}
We include the full Gemini-2.5-Pro prompt used for annotating the human-motion videos in the supplementary materials.

\paragraph{Data Filtering.}
During VLM-based annotation, we instruct Gemini to identify videos depicting static or near-static human motions.
To further remove under-expressive sequences from a kinematic perspective, we apply an additional heuristic: a motion sequence is filtered out if the velocities of all end-effectors remain below 5 cm/s. 
This threshold corresponds to natural micro-movement during human quietly  standing. 
Combining semantic–kinematic filtering, we ensures the dataset for motion head fine-tuning primarily contains expressive motion patterns.

\section{Zero-shot Text-to-Motion Generation}
\begin{figure}[!ht]
    \centering
    \includegraphics[width=1\linewidth]{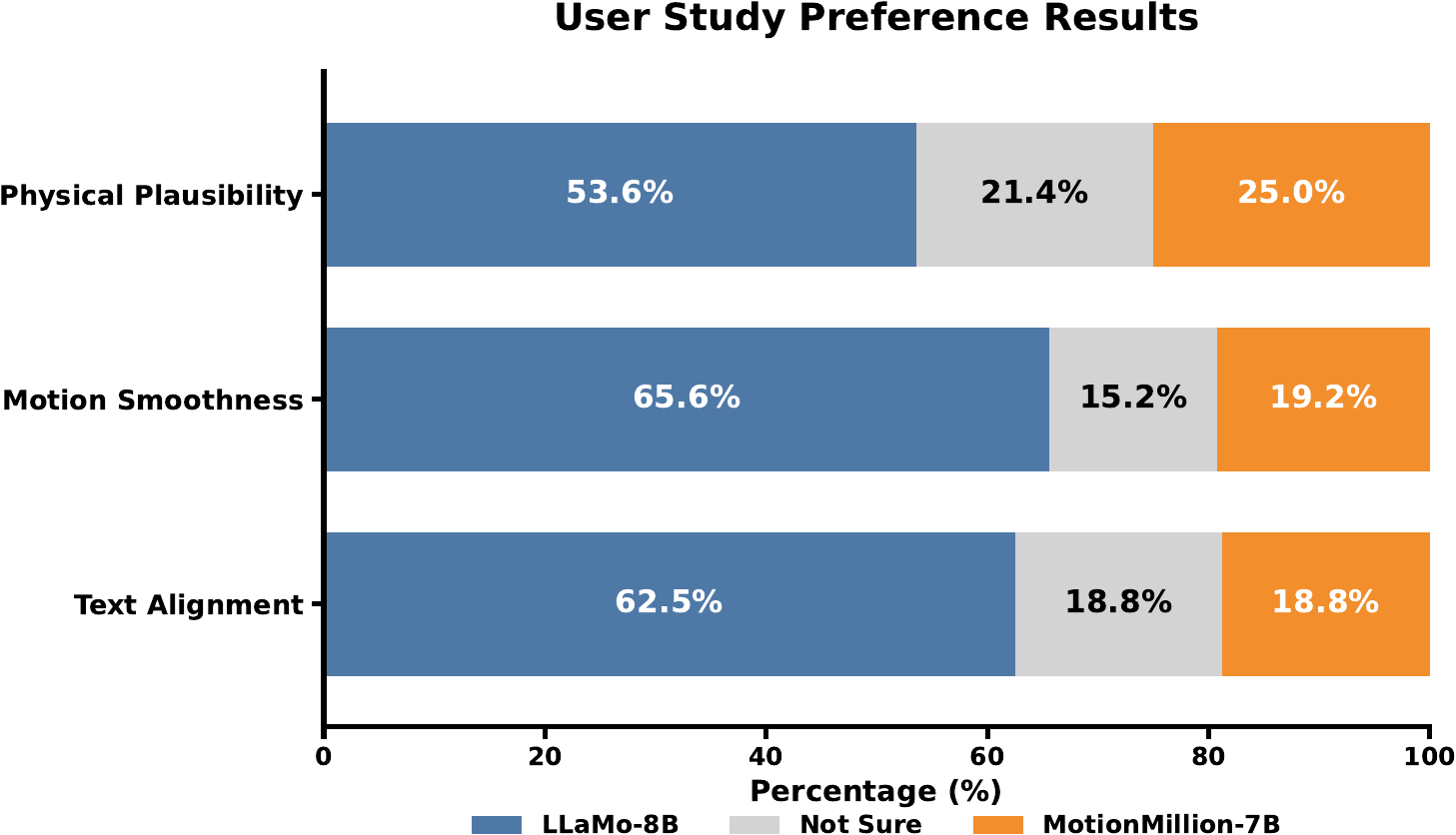}
    \caption{\textbf{User Study of Zero-shot Text-to-Motion Generation.} We use the prompts from MotionMillion-Eval~\cite{fan2025go} to evaluate our model against MotionMillion~\cite{fan2025go}.
    Results show that users significantly prefer our model across all the evaluation axes.}
    \label{fig:user_study}
\end{figure}

In this section, we present additional results demonstrating the zero-shot motion generation capabilities of our large unified model.
\paragraph{User Study versus MotionMillion~\cite{fan2025go}.}
We conducted an A/B human evaluation study with 14 participants to assess motion generation quality across three dimensions: \textbf{Physical Plausibility}, \textbf{Motion Smoothness}, and \textbf{Text Alignment}. 
In this study, participants were shown motions generated by each model for the same text prompt, without knowing which model produced which motion.
As shown in~\cref{fig:user_study}, our model achieves substantially better performance than the current state-of-the-art, MotionMillion~\cite{fan2025go}, across all metrics.
Leveraging high-fidelity continuous motion representations, \name produces noticeably smoother and more physically plausible human motions compared with MotionMillion.
Furthermore, our model demonstrates superior text–motion alignment, even though both MotionMillion and \name employ comparable parameter budgets for text tokens (see~\cref{tab:params_comp}). 
This highlights the effectiveness of our strategy around retaining strong native language capabilities in the underlying LLM while enabling high-quality motion generation.
\begin{table}[!ht]
    \centering
    \resizebox{1\linewidth}{!}{
    \begin{tabular}{lccc}
        \toprule
        Methods & Motion Activated \#Params & Text Activated \#Params & Total \#Params\\
        \midrule
        MotionMillion-3B & 3B & 4.2B & 4.2B \\
        MotionMillion-7B & 7B & 8.2B & 8.2B \\
        \name-1B & 1B & 1B & 2B \\
        \name-3B & 3B & 3B & 6B \\
        \name-8B & 8B & 8B & 16B \\
        \bottomrule
    \end{tabular}}
    \caption{\textbf{Parameters Comparison for Each Modality.} MotionMillion-7B has similar text token activated parameters with \name-8B, which indicates similar language modeling capacity.}
    \label{tab:params_comp}
\end{table}


\paragraph{Generalization to Unseen Languages.}
While studying the zero-shot capabilities of \name, we came across an interesting emergent behavior: \textit{We notice that \name is able to generate motion from prompts in languages beyond English, even though our training data only had English language-motion data.}
We highlight this intriguing emergent behavior as a qualitative observation by showing some examples in the supplementary. Please open the results webpage in our supplementary materials and allow 1-2 minutes for the webpage to load the videos. You can also click on the thumbnails / black tiles if they are not loaded.

\begin{figure}[t]
    \centering
    \begin{subfigure}[b]{\linewidth}
        \centering
        \includegraphics[width=\linewidth]{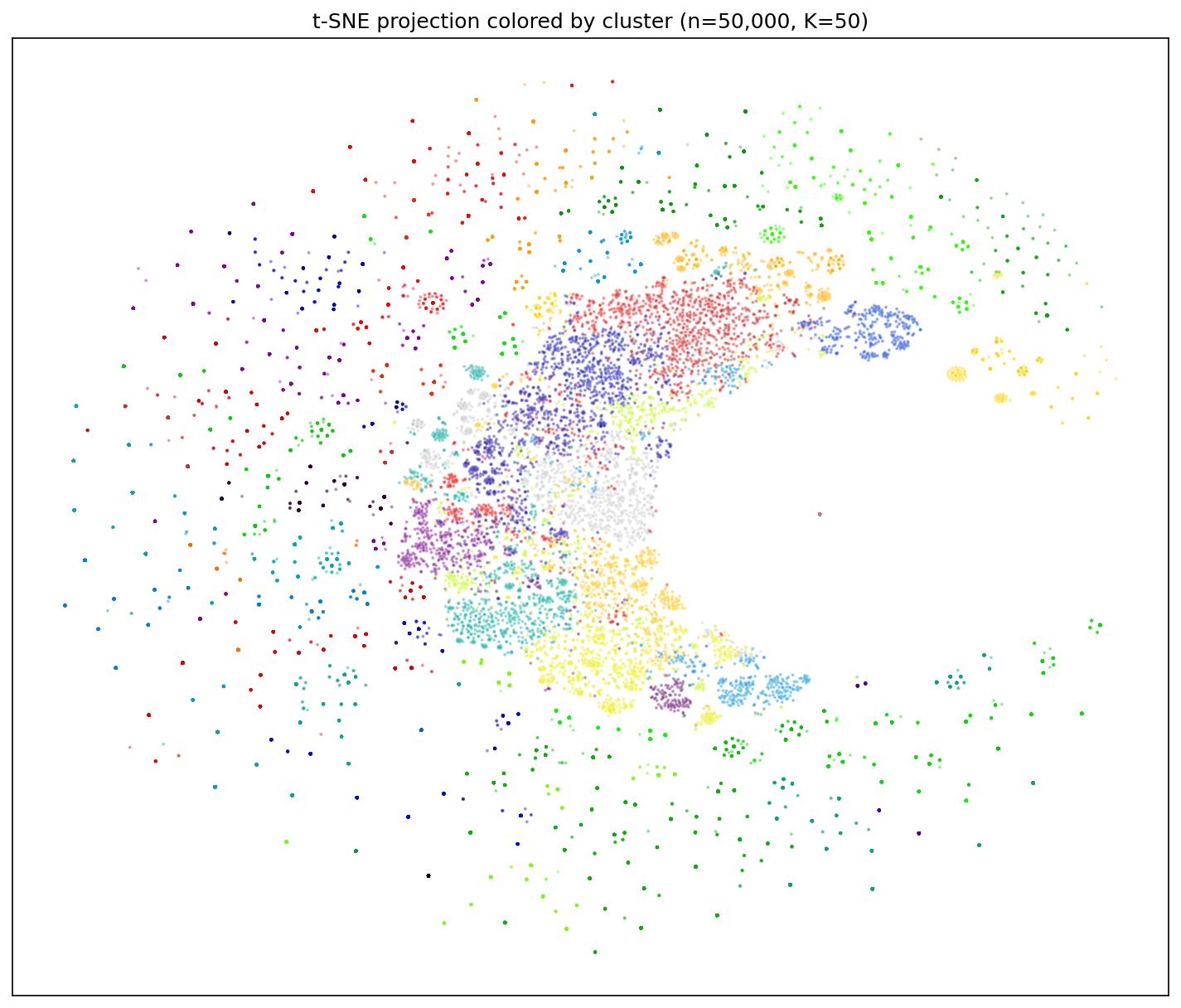}
        \caption{MotionMillion text embedding cluster projection.}
        \label{fig:mm_cluster}
    \end{subfigure}

    \vspace{6pt}  

    \begin{subfigure}[b]{\linewidth}
        \centering
        \includegraphics[width=\linewidth]{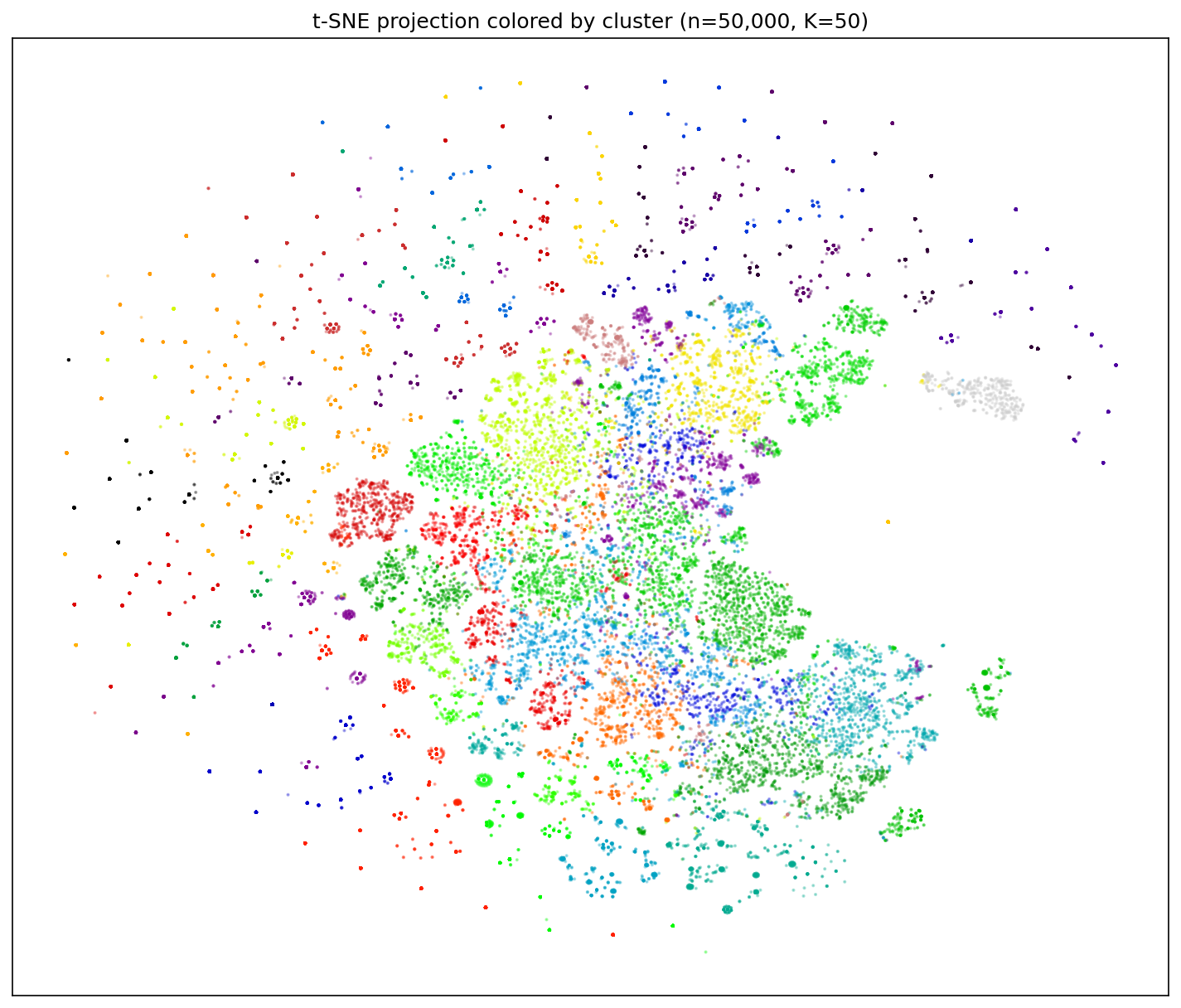}
        \caption{Internal dataset text embedding cluster projection.}
        \label{fig:sstk_cluster}
    \end{subfigure}

    \caption{Semantic distribution visualization. We apply t-SNE to project the learned embeddings from both datasets into a shared 2D space. K-means clustering is performed independently on each dataset. For clarity, we subsample data points according to the density of each cluster.}
    \label{fig:cluster_projection}
\end{figure}

\paragraph{Dataset Comparison}
To demonstrate the semantic limitation about existing open-source large text-motion dataset, we visualize the text embedding distributions based on t-SNE.
To extract the robust embeddings related to the human motion within the massive text annotation, we use Qwen3-Embedding-0.6B for sentence encoding.
The instruction prompt we used is shown as followed. \\
\begin{promptbox}
Instruct: Given a text description of human motion, encode the semantic meaning of the described body movement, action, and style\textbackslash nQuery:
\end{promptbox}
\noindent The t-SNE visualization in~\cref{fig:cluster_projection} reveals that MotionMillion exhibits pronounced inter-cluster gaps and notably smaller intra-cluster spread compared to our internal dataset. 
This suggests that, despite its large scale, MotionMillion suffers from limited semantic diversity: samples are concentrated around a narrow set of motion patterns, resulting in substantial semantic redundancy. 
In contrast, our internal dataset demonstrates more uniform coverage across the embedding space, indicating broader and more balanced motion knowledge.

\begin{figure*}[t]
    \centering
    \begin{subfigure}[b]{\linewidth}
        \centering
        \includegraphics[width=\linewidth]{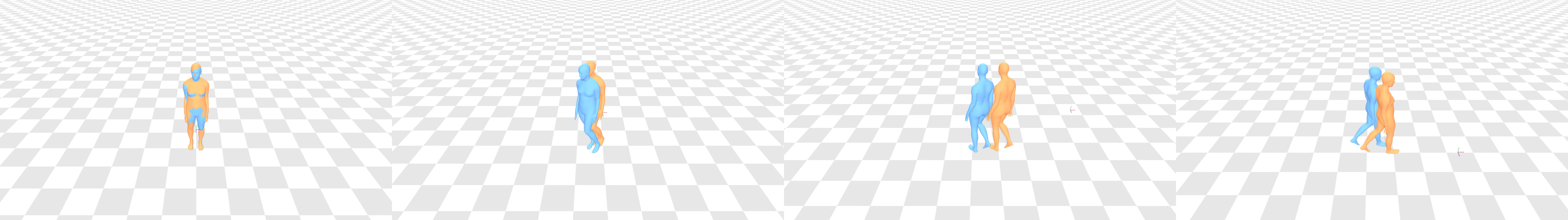}
        \caption{Motion Reconstruction by FSQ in MotionMillion.}
    \end{subfigure}

    \vspace{6pt}  

    \begin{subfigure}[b]{\linewidth}
        \centering
        \includegraphics[width=\linewidth]{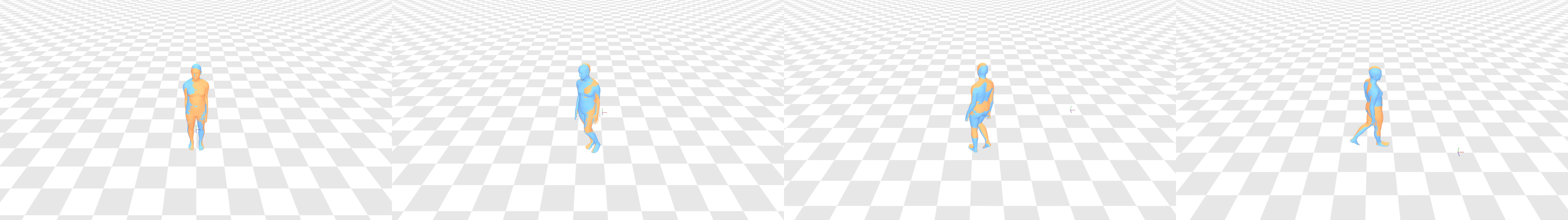}
        \caption{Motion Reconstruction by sigma causal TAE in LLaMo.}
    \end{subfigure}

    \caption{Qualitative Comparison about Motion Reconstruction. The blue motion is ground truth motion and the orange motion is the reconstructed motion.}
    \label{fig:vae_comp}
\end{figure*}

\section{Motion Auto-encoder Comparison}
We present a qualitative comparison between the FSQ tokenizer adopted in MotionMillion~\cite{fan2025go} and the sigma-causal TAE employed in LLaMo. As shown in \cref{fig:vae_comp}, the discrete tokenization in FSQ introduces inherent information loss during the quantization step, resulting in noticeable reconstruction artifacts such as jittery motion and loss of fine-grained details. In contrast, our continuous latent representation preserves higher fidelity to the original motion, enabling more accurate and temporally coherent reconstruction.

\end{document}